\documentclass[10pt,twocolumn,letterpaper,dvips]{article}
\usepackage{cvpr}
\usepackage{times}
\usepackage{epsfig}
\usepackage{graphicx}
\usepackage{amsmath}
\usepackage{amssymb}
\usepackage[breaklinks=true,bookmarks=false]{hyperref}
\cvprfinalcopy 
\def\cvprPaperID{****} 

\begin{document}
%
\title{Spatial Frequency Loss for Learning Convolutional Autoencoders}
\author{Naoyuki Ichimura\\
Human Informatics Research Institute\\
National Institute of Advanced Industrial Science and Technology~(AIST) \\
1-1-1, Umezono, Tsukuba, Ibaraki, 305-8568, JAPAN\\
{\tt\small naoyuki.ichimura@aist.go.jp}
}
\maketitle
%
\begin{abstract}
This paper presents a learning method for convolutional
autoencoders~(CAEs) for extracting features from images. CAEs can be
obtained by utilizing convolutional neural networks to learn an
approximation to the identity function in an unsupervised manner. The
loss function based on the pixel loss~(PL) that is the mean squared
error between the pixel values of original and reconstructed images is
the common choice for learning. However, using the loss function leads
to blurred reconstructed images. A method for learning CAEs using a
loss function computed from features reflecting spatial frequencies is
proposed to mitigate the problem. The blurs in reconstructed images
show lack of high spatial frequency components mainly constituting
edges and detailed textures that are important features for tasks such
as object detection and spatial matching. In order to evaluate the
lack of components, a convolutional layer with a Laplacian filter bank
as weights is added to CAEs and the mean squared error of features in
a subband, called the spatial frequency loss~(SFL), is computed from
the outputs of each filter. The learning is performed using a loss
function based on the SFL. Empirical evaluation demonstrates that
using the SFL reduces the blurs in reconstructed images.
\end{abstract}
%
%
%
\section{Introduction}
Extracting features from images is crucial as a basis for tasks such
as object detection and spatial matching. Although there are several
feature extraction methods, extracting features using deep learning
has attracted much attention. Using deep learning, we can
automatically construct a feature extraction algorithm as a mapping
represented by a multi-layer neural network. In particular,
convolutional neural networks~(CNNs)\cite{LeCun90} in which filters
for convolutional operations are automatically learnt have been widely
used. The typical example of using CNNs is category classification for
images. CNNs learnt using a large-scale labeled image dataset can
extract features from a lot of
objects\cite{Krizhevsky12,Simonyan15}. The features can even be
transferred to tasks other than classification. However, annotating a
large number of images is time-consuming, thus considering feature
extraction based on unsupervised learning is necessary.\par
A way to realize feature extraction based on unsupervised learning is
to use autoencoders\cite{Salakhutdinov06}. Autoencoders can be
obtained by utilizing multi-layer neural networks to learn an
approximation to the identity function in an unsupervised
manner. Several investigators have shown that the features obtained
from hidden layers of autoencoders are useful for the tasks such as
image retrieval\cite{Krizhevsky11}, image classification\cite{Le12}
and image clustering\cite{Xie16}. The architecture of the networks
used in \cite{Krizhevsky11} and \cite{Xie16} are fully connected, and
the network used in \cite{Le12} has local receptive fields that are
not convolutional. Although the networks are reasonable to approximate
the identity function, images with the same resolution have to be fed
into the networks.\par
A learning algorithm for convolutional autoencoders~(CAEs) for
extracting features from images is considered in this paper. In CAEs,
the inputs and outputs of CNNs are original and reconstructed images,
and the differences between them are tried to minimize to learn an
approximation to the identity function. The following advantages are
main reasons to choose CAEs:(1) images with arbitrary resolutions can
be used both in learning and prediction phases owing to weight
sharing, (2) spatial dependency of features can be controlled by
changing the resolutions of feature maps in hidden layers, (3)
connecting CAEs to other CNNs is easy. Beside these advantages, CAEs
have a well-known issue: learning CAEs using the loss function based
on the pixel loss~(PL) that is the mean squared error between the
pixel values of original images and reconstructed images leads to
blurred reconstructed images. Since the blurs show lack of information
in features extracted in hidden layers to reproduce original images,
to address the issue is important for feature extraction.\par
Another way for feature extraction via unsupervised learning is to use
deep convolutional generative adversarial
networks~(DCGANs)\cite{Goodfellow14,Makhzani15,Denton15,Radford16}. Two
CNNs called a generator and a discriminator are used in DCGANs. Images
are generated from random variables in the former, discrimination
between generated images and corresponding original images is
performed in the latter. The objective of learning DCGANs is to
generate images that cannot be discriminated from corresponding
original images. It's shown that the images generated by DCGANs have
little blurs. The reduction of blurs comes from the evaluation of
generated images not using pixel values but using features utilizing
in a discriminator. It confirms that the evaluation using features
extracted by networks is useful in image style transfer as
well\cite{Gatys16,Johnson16}.\par
A way of addressing the issue of CAEs is considered here while keeping
the image evaluation using features obtained by networks in mind. The
blurs in reconstructed images show lack of high spatial frequency
components mainly constituting edges and detailed textures that are
important features for tasks such as object detection and spatial
matching. That is, learning based on the PL does not exploit the
information on high spatial frequency components. In order to exploit
the information, a method for learning CAEs using a loss function
computed from features reflecting spatial frequencies is proposed. The
proposed method extracts features reflecting spatial frequencies from
both original and reconstructed images by a Laplacian filter bank that
has band-pass property. More precisely, a convolutional layer with a
Laplacian filter bank as weights is added to CAEs and the mean squared
error of features in a subband, called the spatial frequency
loss~(SFL), is computed from the outputs of each filter. Since we can
separate features with high spatial frequencies from features with low
spatial frequencies by a Laplacian filter bank, the learning algorithm
based on the SFL exploits the information on high spatial frequency
components to reproduce original images.\par
The differences between the proposed method and DCGANs are twofold:
(1)~The evaluation of images in DCGANs is performed using features
obtained by a discriminator. In this case, the information reflected
by the features is determined by an image dataset used in
learning. Thus it's not ambiguous that the features have the
information on specific spatial frequencies. On the other hand, the
proposed method uses the features explicitly reflecting specific
spatial frequencies to evaluate images. (2)~Since only a single layer
has to be added to CAEs to extract the features by a Laplacian filter
bank, the additional computational costs in learning is quite small
compared to DCGANs. As demonstrated in the experimental results,
introducing the features reflecting specific spatial frequencies is
clearly effective to reduce the blurs in reconstructed images.\par
This paper organized as follows: Section \ref{sec:CAEs} explains
CAEs. The learning method using the loss function based on the SFL is
proposed in Sec.\ref{sec:learningCAEs}. The experimental results are
shown in Sec.\ref{sec:exp_results}. Section \ref{sec:summary}
summarizes this paper.
%
%
%
\section{Convolutional Autoencoders~(CAEs)}
\label{sec:CAEs}
%
%
\begin{figure}[t]
\begin{center}
\includegraphics[width=8.0cm]{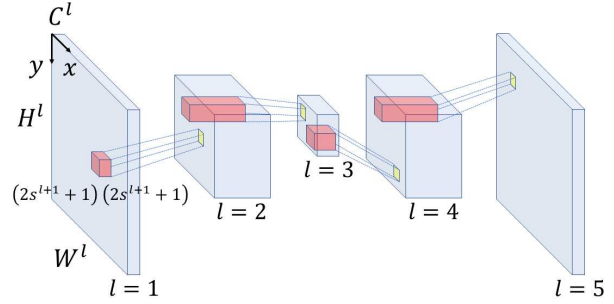}
\end{center}
\vspace*{-0.2cm}
\caption {
An example of a CAE. The number of layer $L$ is five. A mapping
between the input layer and the output layer is realized by applying
the filters with the size
$C^{l+1}\left(2s^{l+1}+1\right)\left(2s^{l+1}+1\right)$ to the volume data
with the size $C^lW^lH^l$. The output layer has feature maps with the
same size as an original image and learning is performed to reproduce
the original image in the feature maps.
}
\label{fig:CAE_example}
\vspace*{-0.2cm}
\hrulefill
\vspace*{-0.3cm}
\end{figure}
A fully convolutional network\cite{Springenberg15,Shelhamer17} is used
to constitute a CAE in this paper. It assumes that a $L$ layer CNN
such as Fig.\ref{fig:CAE_example} has the volume data with the number
of channels $C^l$, width $W^l$ and height $H^l$, $l=1,\ldots,L$. The
volume data are represented as follows:
\begin{align}
&f^l\left(c,x,y\right),
\label{eq:fmap}\\
&c=1,\ldots,C^l,~x=1,\ldots,W^l,~y=1,\ldots,H^l\notag.
\end{align}
The voxels in $f^l\left(c,x,y\right)$ correspond to artificial
neurons. The volume data in the $\left(l+1\right)$\--th layer are
computed from one in the $l$\--th layer using convolution operations
and an activation function as follows:
\begin{align}
&\hspace*{-0.2cm}o^{l+1}\left(c,x,y\right)=\sum_{c'=1}^{C^l}\cdot\notag\\
&\hspace*{-0.3cm}\sum_{\alpha=-s^{l+1}}^{s^{l+1}}\sum_{\beta=-s^{l+1}}^{s^{l+1}}
  f^l\left(c',x+\alpha,y+\beta\right)w^{l+1}(c,c',\alpha,\beta)
\label{eq:convolution}\\
&\hspace*{-0.2cm}f^{l+1}\left(c,x,y\right) = a^{l+1}\left(o^{l+1}\left(c,x,y\right)+b^{l+1}\left(c\right)\right)
\label{eq:activation}
\end{align}
where, $w^{l+1}(c,c',\alpha,\beta)$,~$c=1,\ldots,C^{l+1}$ are filters
with the size $C^{l+1}\left(2s^{l+1}+1\right)\left(2s^{l+1}+1\right)$,
$s^{l+1}\in\mathcal{N}$, $b^{l+1}\left(c\right)$ are biases and
$a^{l+1}\left(\cdot\right)$ is an activation function. The variables
$\alpha$ and $\beta$ denote the horizontal and vertical shifts from
$\left(x,y\right)$ and their origin is the center of a filter. Since
the equations (\ref{eq:convolution}) and (\ref{eq:activation}) can be
regarded as feature extraction by convolution operations,
$f^l\left(c,x,y\right)$,~$l>1$ are called feature maps.\par
The input layer of a CAE is an original image and the output layer has
feature maps with the same size as an original image. The filters and
biases of a CAE are adjusted by learning so as to reproduce the
original image in the feature maps of the output layer. A five layer
CAE is shown in Fig.\ref{fig:CAE_example} as an example.\par
The number of channels of hidden layers of CAEs should be smaller than
the total number of channels of original images in an image dataset
used in learning so that CAEs tries to reconstruct original images
from the restricted features appeared in hidden layers. It turns out
to extract common and useful features from original images to
reconstruct them. Since no annotating images is required, the feature
extraction by CAEs is realized by unsupervised learning.
%
%
%
\section{Learning CAEs Using Spatial Frequency Loss}
\label{sec:learningCAEs}
%
%
\begin{figure}[t]
\begin{center}
\includegraphics[width=8.0cm]{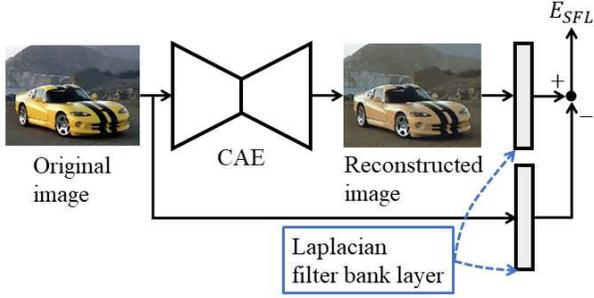}
\end{center}
\vspace*{-0.1cm}
\caption {
 An example of the network for learning a CAE using the loss function
 $E_{SFL}$ based on the proposed spatial frequency loss. A
 convolutional layer with the weights obtained from a Laplacian filter
 bank is added to the CAE. The loss function $E_{SFL}$ is computed by
 inputting original and reconstructed images to the Laplacian filter
 bank layer to extract features reflecting spatial frequencies.
}
\label{fig:CAE_with_sfl}
\vspace*{-0.2cm}
\hrulefill
\vspace*{-0.3cm}
\end{figure}
This section proposes a novel loss function for learning CAEs. We can
use any loss functions that are consistent with the purpose of
reconstructing original images. One of the loss functions is the
following $E_{PL}$ based on the PL that is the mean squared error
between the pixel values of original and reconstructed images:
\begin{align}
&\hspace*{-0.2cm}E_{PL}=\frac{1}{N}\sum_{i=1}^N\sum_{c=1}^{c^L}w_{PL}\left(c\right)E_{PL}^{L}\left(i,c\right)
\label{eq:pl}\\
&\hspace*{-0.2cm}E_{PL}^{L}\left(i,c\right)=\frac{1}{2}\frac{1}{W^LH^L}\cdot
\notag\\
&\hspace*{0.3cm}\sum_{x=1}^{W^L}\sum_{y=1}^{H^L}\left\{f_i^L\left(c,x,y\right)-f_i^1\left(c,x,y\right)\right\}^2
\notag
\end{align}
where, $N$ is the number of images in a dataset,
$w_{PL}\left(c\right)$ are the weights for the channels in the
$L$\--th layer and the subscript $i$ for the volume data represents
the $i$\--th image in a dataset. Although $E_{PL}$ is straightforward
for the purposed of CAEs, learning using $E_{PL}$ leads to blurred
reconstructed images, which show lack of high spatial frequency
components.\par
High spatial frequency components mainly constitute edges and detailed
textures that are important features for tasks such as object
detection and spatial matching. Thus generating more clear
reconstructed images by compensating the lack of components is
important for feature extraction. In order to generate more clear
reconstructed images, a learning method for CAEs using a loss function
computed from features reflecting spatial frequencies is proposed. In
the proposed method, a convolutional layer with a Laplacian filter
back as weights is added to CAEs, and the loss function is computed
from the outputs of the layer. Figure \ref{fig:CAE_with_sfl} shows an
example of the network for learning a CAE using the proposed loss
function.\par
%
%
%
\begin{figure}[t]
\begin{center}
\includegraphics[width=8.0cm]{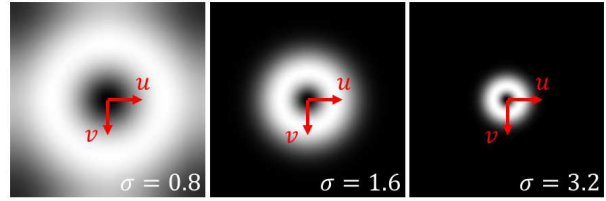}
\end{center}
\vspace*{-0.1cm}
\caption {
The frequency responses of the Laplacian filters with the scales
$\sigma=0.8,~1.6$ and $3.2$. The $u$ and $v$ in these figures
represent the spatial frequencies for $x$ and $y$ directions,
respectively. The brightness of the images show the power spectra. As
we can see from the figures, the Laplacian filter has band-pass
property.
}
\label{fig:NLoG_fr}
\vspace*{-0.2cm}
\hrulefill
\vspace*{-0.3cm}
\end{figure}
\begin{figure}[!t]
\begin{center}
\includegraphics[width=3.8cm]{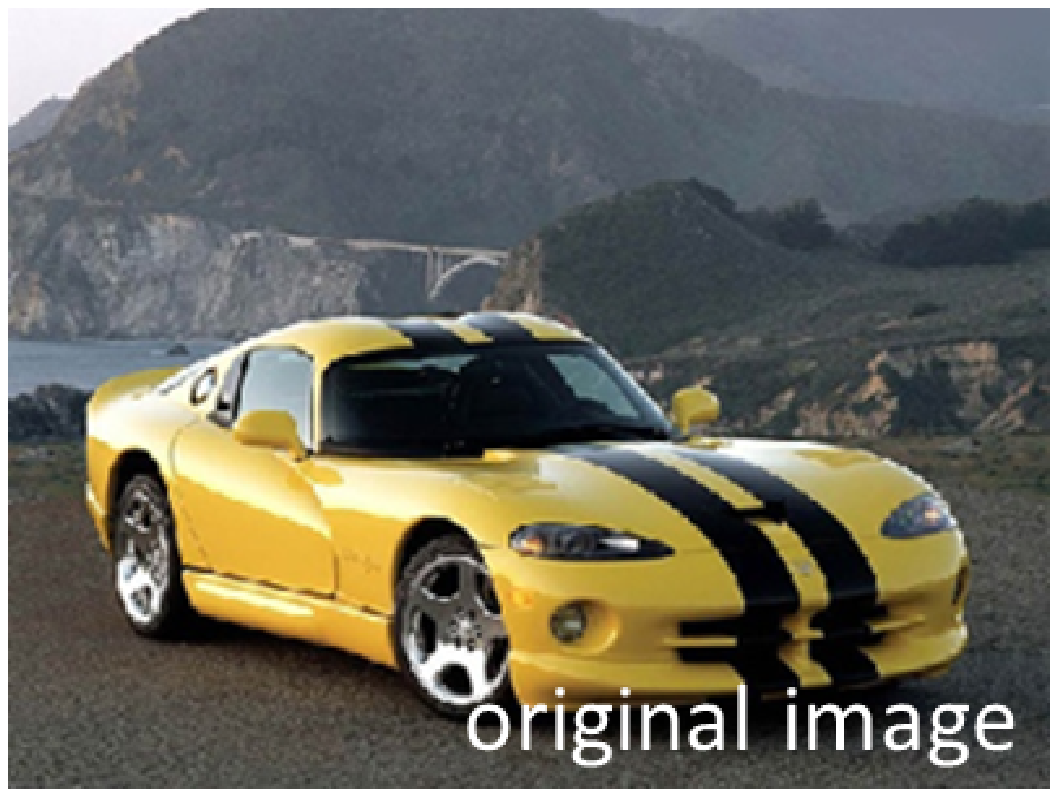}
\includegraphics[width=3.8cm]{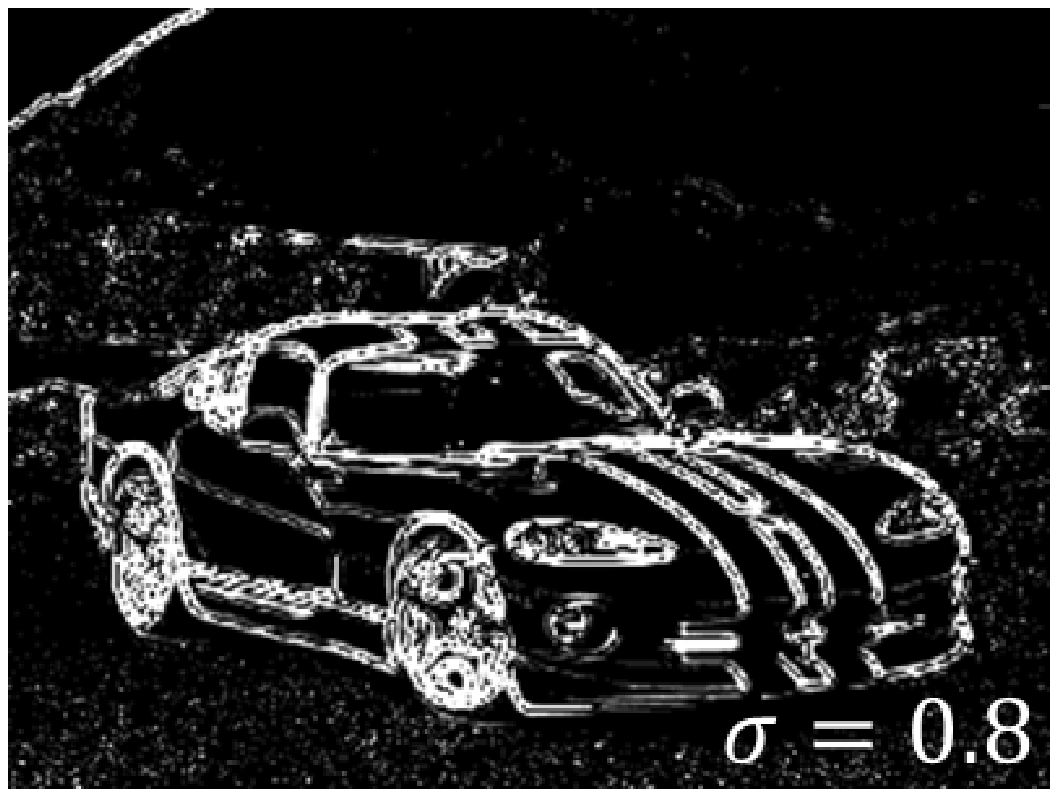}\\
\includegraphics[width=3.8cm]{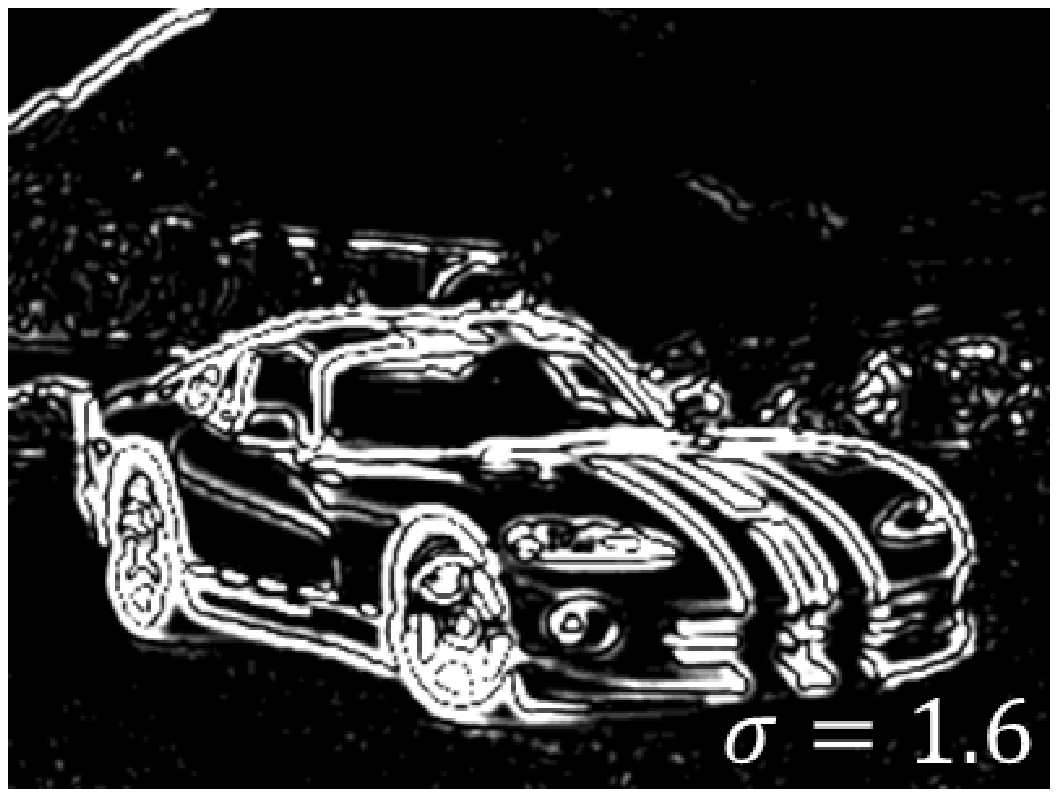}
\includegraphics[width=3.8cm]{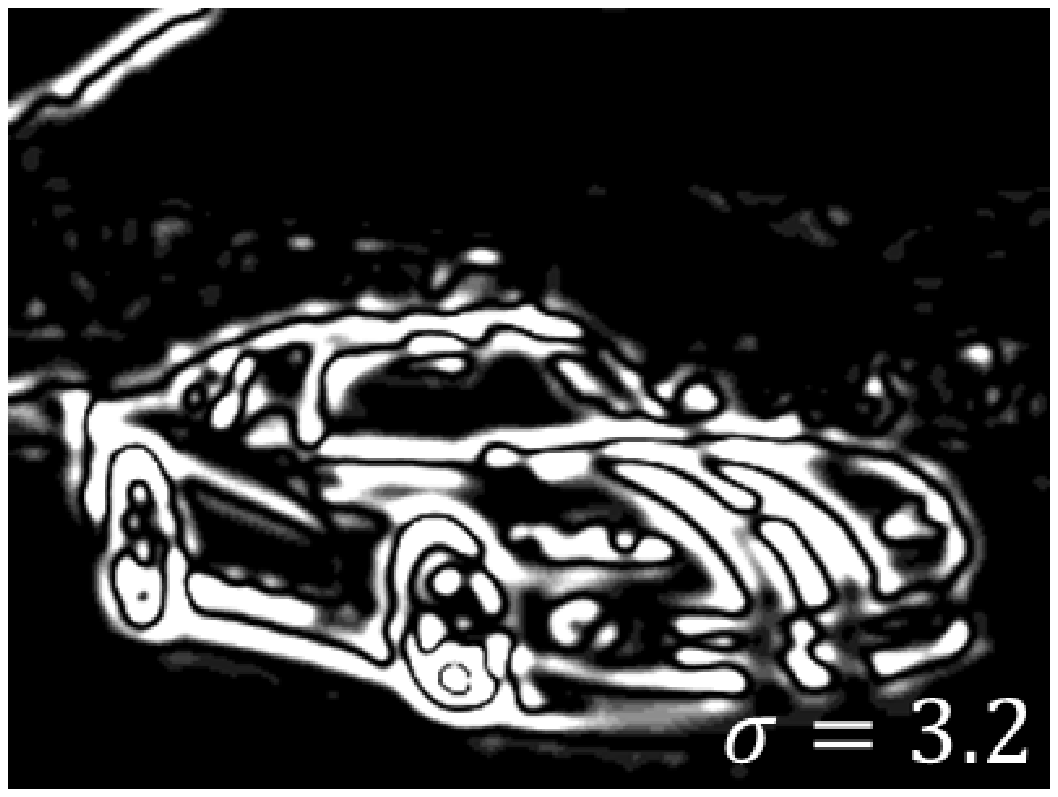}
\end{center}
\vspace*{-0.1cm}
\caption {
An example of the outputs of the Laplacian filter bank. The outputs
are squared and quantized to the uniform 10 levels in [0:255]. The outputs
reflect the spatial frequencies because the smaller scale is used, the
smaller changes in brightness are extracted.
}
\label{fig:NLoG_filterbank_example}
\vspace*{-0.2cm}
\hrulefill
\vspace*{-0.3cm}
\end{figure}

The Laplacian filter has the coefficients obtained from the following
normalized Laplacian of Gaussian function:
\begin{align}
&\sigma^2\nabla^2g\left(x,y\right)=-\left(2-\frac{x^2+y^2}{\sigma^2}\right)g\left(x,y\right)
\label{eq:nlog}\\
&g\left(x,y\right)=\frac{1}{2\pi\sigma^2}e^{-\frac{x^2+y^2}{2\sigma^2}}
\notag
\end{align}
where, $\sigma$ is the scale. Figure \ref{fig:NLoG_fr} shows the
frequency responses of the Laplacian filters with the scales
$\sigma=0.8,1.6$ and $3.2$. As we can see from these figures, the
Laplacian filter has band-pass property. The subband passed by the
filter varies with the scale; the smaller scale is used, the higher
spatial frequency is passed. Thus using the Laplacian filter bank, we
can extract features in each subband from original and reconstructed
images. Figure \ref{fig:NLoG_filterbank_example} represents an example
of the outputs of the Laplacian filter bank. The outputs reflect the
spatial frequencies because the smaller scale is used, the smaller
changes in brightness are extracted.\par
%
%
The definition of the spatial frequency loss~(SFL) is now shown. Let a
convolutional layer added to CAEs be the $\left(L+1\right)$\--th
layer. The activation function of the $\left(L+1\right)$\--th layer is
the identity function and the biases are 0. The SFL denoted by
$E_{SFL}^{L+1}$ is defined by the mean squared error of features in a
subband computed from the outputs of the $\left(L+1\right)$\--th
layer, and the loss function $E_{SFL}$ is constructed from the SFL as
follows:
\begin{align}
&\hspace*{-0.2cm}E_{SFL}=\frac{1}{N}\sum_{i=1}^N\sum_{c=1}^{C^{L+1}}w_{SFL}\left(c\right)E_{SFL}^{L+1}\left(i,c\left|\sigma_c\right.\right)
\label{eq:sfl}\\
&\hspace*{-0.2cm}E_{SFL}^{L+1}\left(i,c\left|\sigma_c\right.\right)=\frac{1}{2}\frac{1}{W^{L+1}H^{L+1}}\cdot
\notag\\
&\hspace*{0.1cm}\sum_{x=1}^{W^{L+1}}\sum_{y=1}^{H^{L+1}}\left\{f_i^{L+1}\left(c,x,y\left|\sigma_c\right.\right)-f_i^{1}\left(c,x,y\left|\sigma_c\right.\right)\right\}^2
\notag
\end{align}
where, $w_{SFL}\left(c\right)$ are the weights for subbands and
$f_i^{L+1}\left(c,x,y\left|\sigma_c\right.\right)$ and
$f_i^1\left(c,x,y\left|\sigma_c\right.\right)$ are the results of
applying the Laplacian filter with the scales $\sigma_c$ to
reconstructed and original images, respectively. The number of
channels $C^{L+1}$ is determined by the number of filters in a
Laplacian filter bank.\par
We can evaluate the losses of subbands by $E_{SFL}$. However, the
losses of whole spatial frequencies can not be computed due to the
band-pass property of the Laplacian filter. Therefore, the following
loss function $E$ is used for learning CAEs, so that the losses of
whole spatial frequencies are evaluated by $E_{PL}$.
\begin{equation}
E=E_{PL}+E_{SFL}
\label{eq:loss_function}
\end{equation}
The backpropagation algorithm\cite{Rumelhart86} is applied to learning
CAEs with the loss function of Eq.(\ref{eq:loss_function}). First, the
following gradients of the $E_{SFL}$ with respect to the outputs of
the $\left(L+1\right)$\--th layer are computed in learning:
\begin{align}
&\hspace*{-0.5cm}\frac{\partial E_{SFL}}{\partial o_i^{L+1}\left(c,x,y\left|\sigma_c\right.\right)}=\frac{1}{N_m}\sum_{i=1}^{N_m}\biggl[w_{SFL}\left(c\right)\label{eq:sfl_grad}\cdot\\
&\hspace*{-0.3cm}\left.\frac{1}{W^{L+1}H^{L+1}}\left\{f_i^{L+1}\left(c,x,y\left|\sigma_c\right.\right)-f_i^{1}\left(c,x,y\left|\sigma_c\right.\right)\right\}\right]\notag
\end{align}
where, $N_m$ is the number of images in a mini-batch and
$o_i^{L+1}\left(c,x,y\left|\sigma_c\right.\right)$ are the outputs of
the $\left(L+1\right)$\--th layer obtained by
Eq.(\ref{eq:convolution}) for the $i$\--th image in a mini-batch. The
gradients are backpropagated to the $L$\--th layer. Then, the
following gradients of $E_{PL}$ are computed at the $L$\--th layer:
\begin{align}
&\hspace*{-0.5cm}\frac{\partial E_{PL}}{\partial o_i^{L}\left(c,x,y\right)}=\frac{1}{N_m}\sum_{i=1}^{N_m}\biggl[w_{PL}\left(c\right)\label{eq:pl_grad}\cdot\\
&\hspace*{-0.3cm}\left.\frac{1}{W^{L}H^{L}}\left\{f_i^{L}\left(c,x,y\right)-f_i^{1}\left(c,x,y\right)\right\}
\cdot\frac{\partial f_i^{L}\left(c,x,y\right)}{\partial o_i^{L}\left(c,x,y\right)}\right]\notag
\end{align}
where, $\frac{\partial f_i^{L}\left(c,x,y\right)}{\partial
  o_i^{L}\left(c,x,y\right)}$ are derived from the activation function
$a^L\left(\cdot\right)$. The gradients of Eq.(\ref{eq:pl_grad}) are
added to the gradients backpropagated from the $\left(L+1\right)$\--th
layer, followed by backpropagating the results to hidden layers of
CAEs to adjust filters and biases. The learning algorithm taking both
the PL and SFL into account is performed in this way.
%
%
%
\section{Experimental Results}
\label{sec:exp_results}
%
%
%
\begin{table}
\caption{
The architecture of the CNN for learning. From the first to the fifth
layers constituted a CAE. The sixth layer had the weights obtained
from the Laplacian filter bank in which three filters were used.
}
\vspace*{-0.1cm}
\begin{center}
\begin{tabular}{|c|c|c|c|} 
\hline
Layer        & Filter                  & Stride    & Activation\\
no.          & size                    &           & function \\
\hline
1            &  n/a                    & n/a       & n/a \\
2            &  $32\times3\times3$     & 1         & ReLU \\
3            &  $16\times3\times3$     & 2         & ReLU \\
4            &  $32\times3\times3$     & 0.5       & ReLU \\
5            &  $3\times3\times3$      & 1         & $\tanh$ \\
\hline
6            &  $3\times\lceil 8\sigma_c\rceil\times\lceil 8\sigma_c\rceil$  & 1  & identity \\
\hline
\end{tabular}
\end{center}
\vspace*{-0.5cm}
\label{tab:CNN_architecture}
\vspace*{0.5cm}
\hrulefill
%
%
\end{table}
%
%
%
\begin{figure*}[t]
\begin{center}
\includegraphics[width=7.5cm]{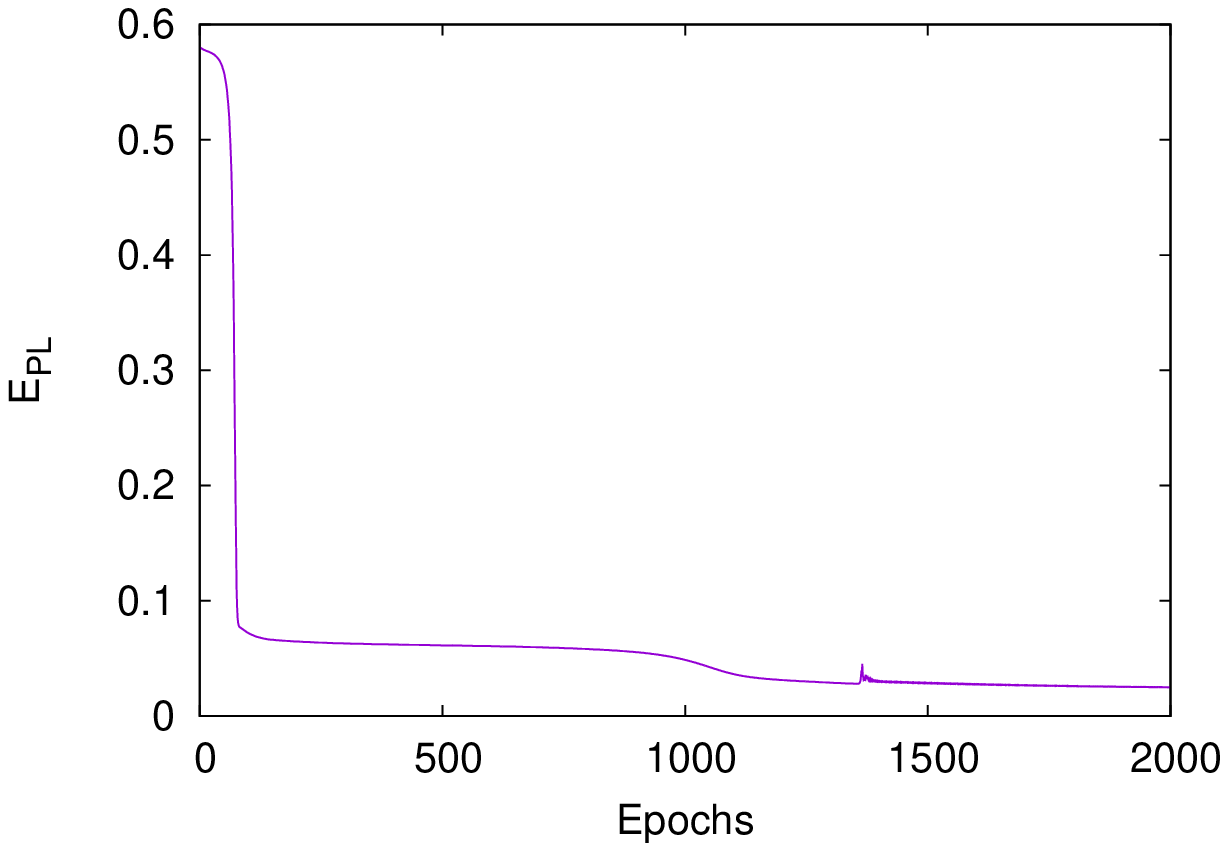}
\includegraphics[width=7.5cm]{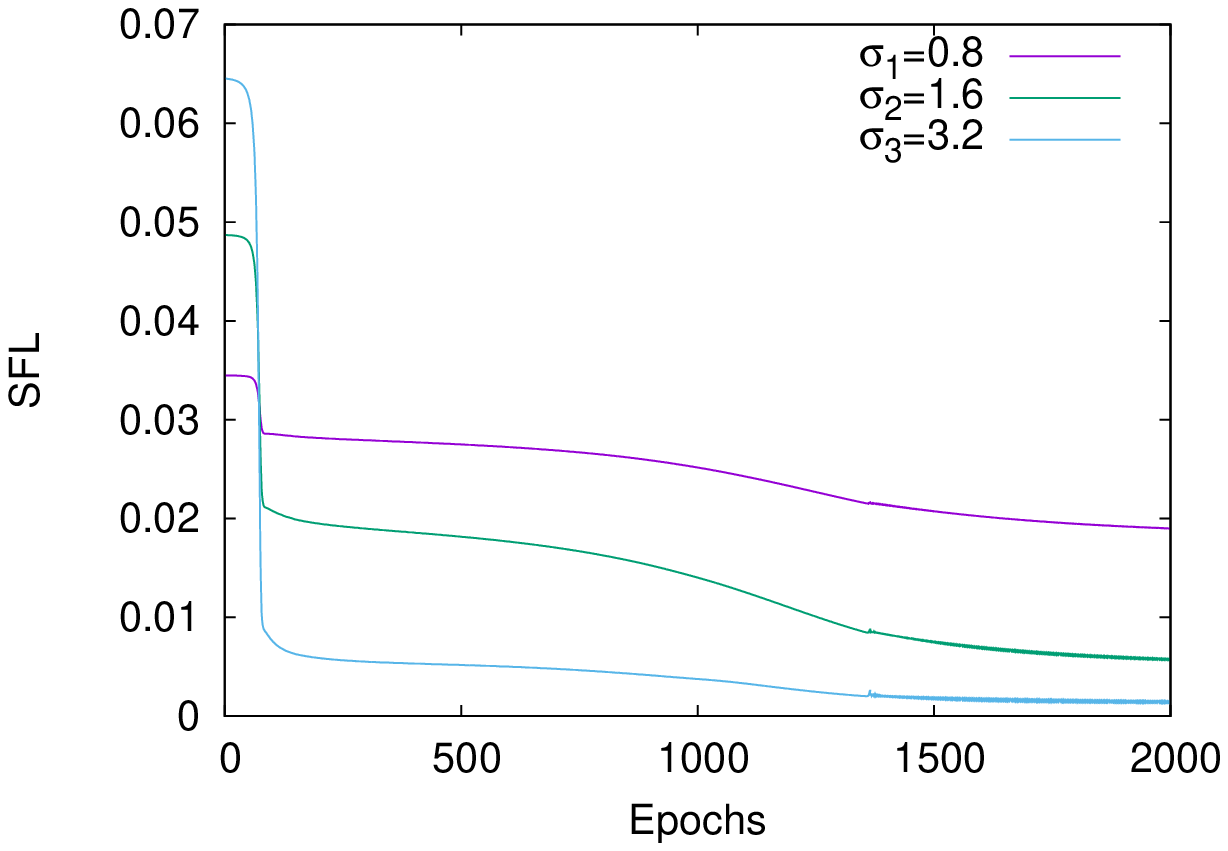}
\end{center}
\vspace*{-0.3cm}
\hspace*{5.1cm}(a)\hspace*{7.3cm}(b)
\vspace*{-0.2cm}
\begin{center}
\includegraphics[width=7.5cm]{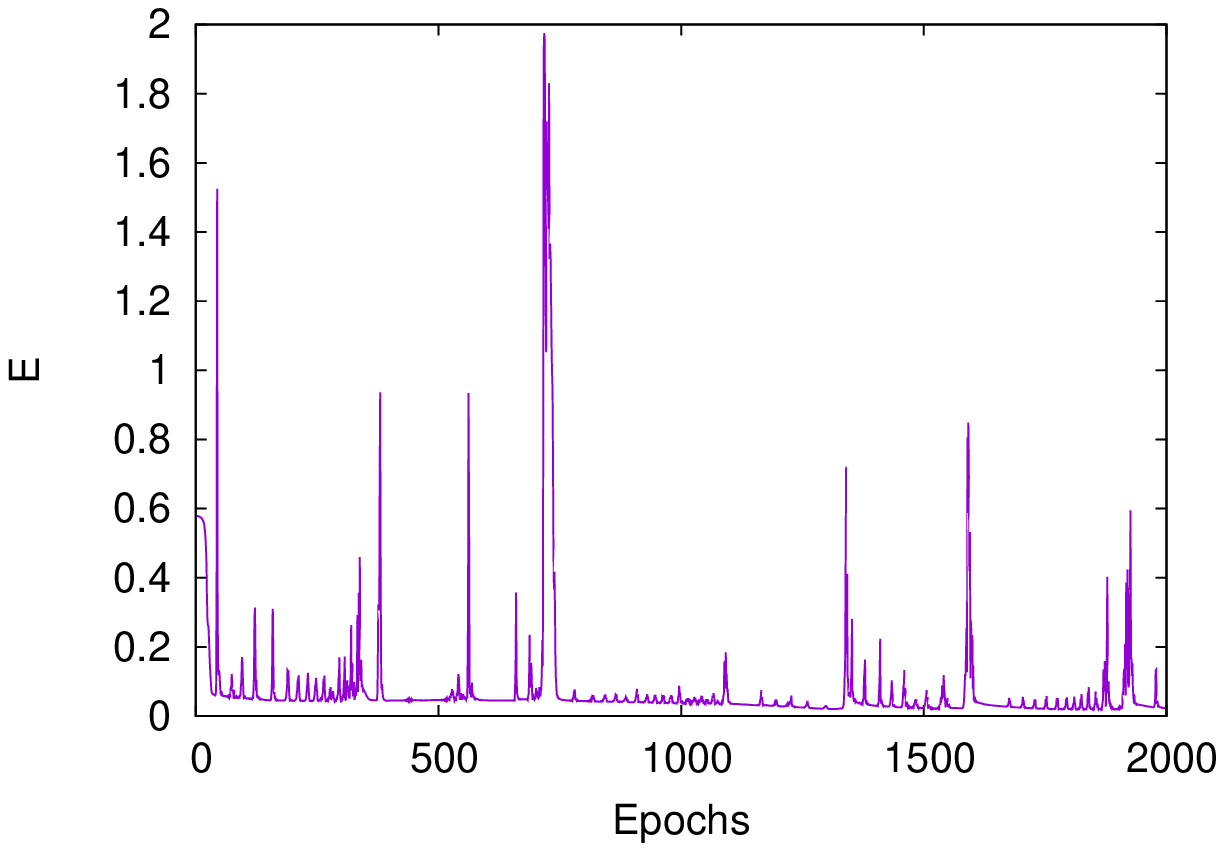}
\includegraphics[width=7.5cm]{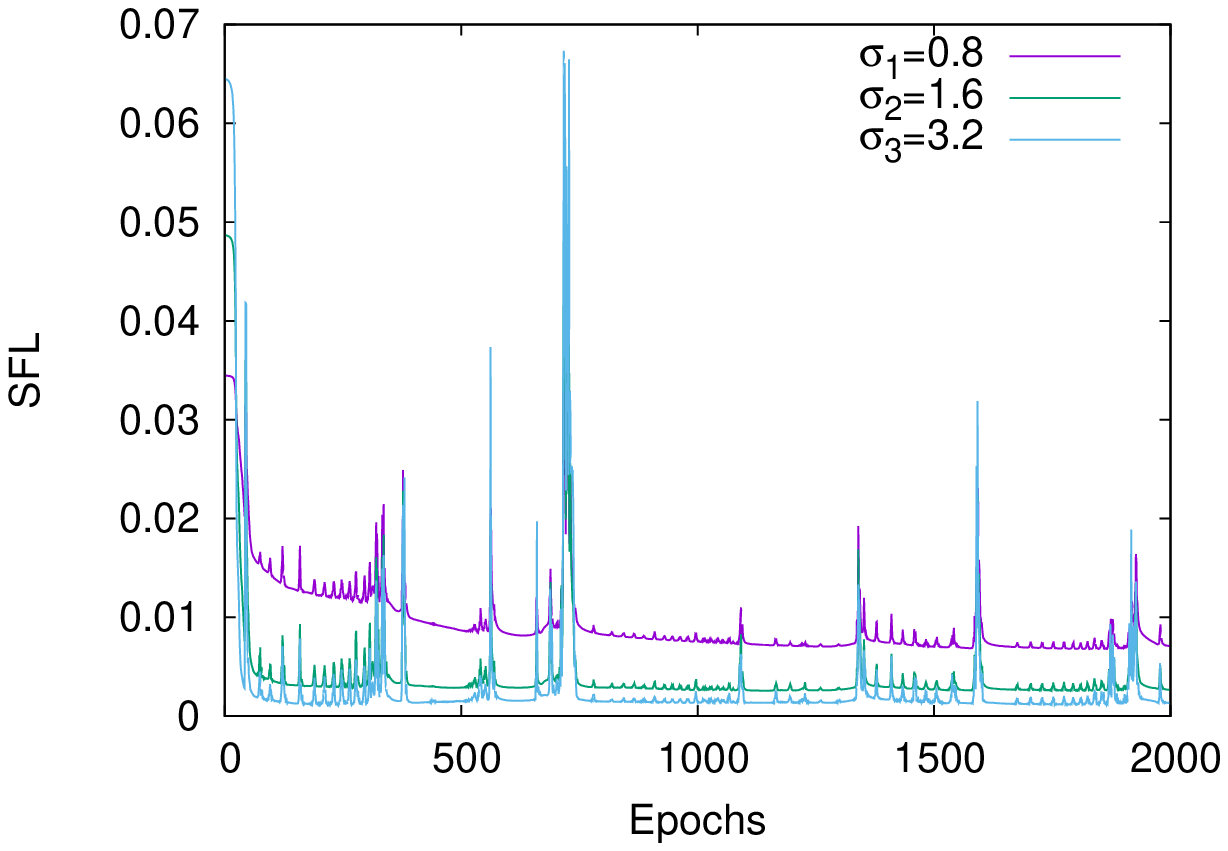}
\end{center}
\vspace*{-0.25cm}
\hspace*{5.1cm}(c)\hspace*{7.3cm}(d)
\vspace*{0.15cm}
\caption {
Learning curves. (a)~The learning using $E_{PL}$. (b)~The changes in
the SFLs of the subbands in learning of (a). (c)~The learning
introducing $E_{SFL}$. (d)~The changes in the SFLs of the subbands in
learning of (c). In (c) and (d), the curves were drawn under the
condition of $W_{SFL}\left(c\right)=1$ to make comparison with (a) and
(b) easy. In learning using $E_{PL}$, the SFLs of the subbands with
$\sigma_c=0.8$ and $1.6$ largely remained even though $E_{PL}$ became
small. On the other hand, the SFLs were reduced by introducing
$E_{SFL}$.
}
\label{fig:learning_curves}
\vspace*{-0.2cm}
\hrulefill
\vspace*{-0.3cm}
\end{figure*}
%
%
%
\begin{figure*}[!t]
%
%
\begin{center}
\includegraphics[height=2.925cm]{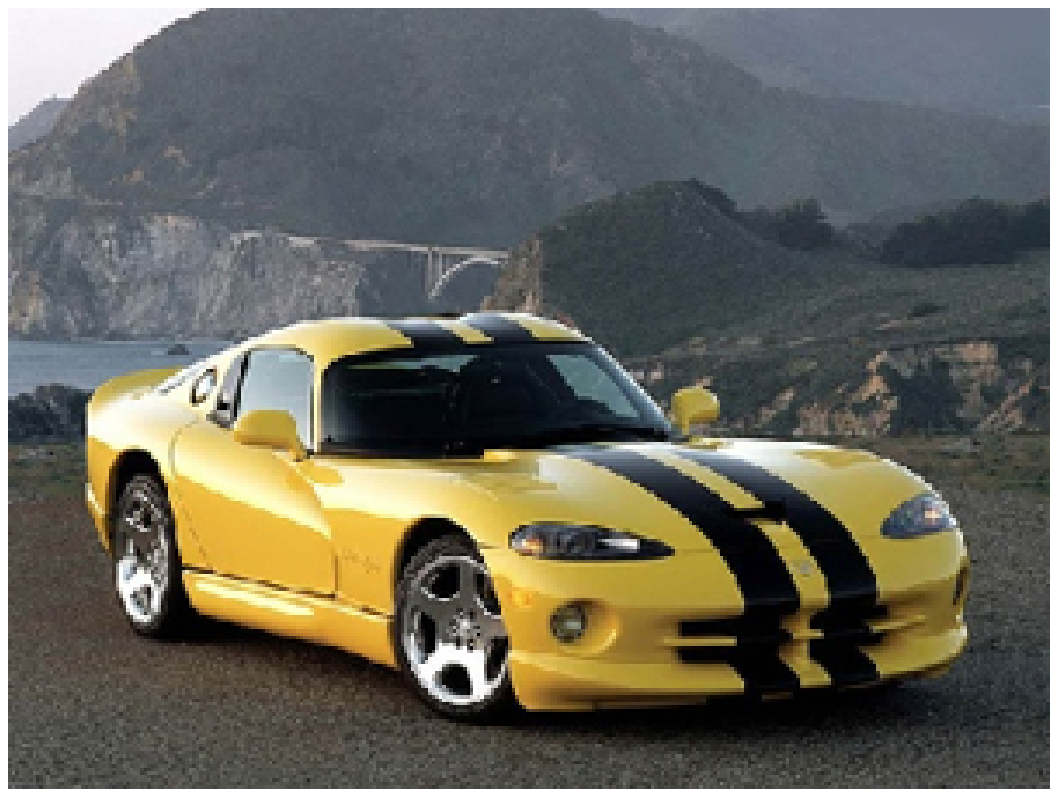}
\includegraphics[height=2.925cm]{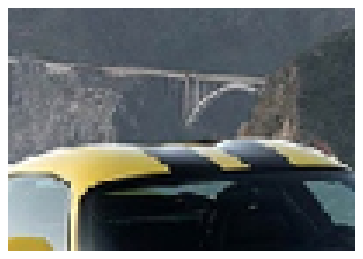}
\hspace*{0.3cm}
\includegraphics[width=2.925cm,angle=90]{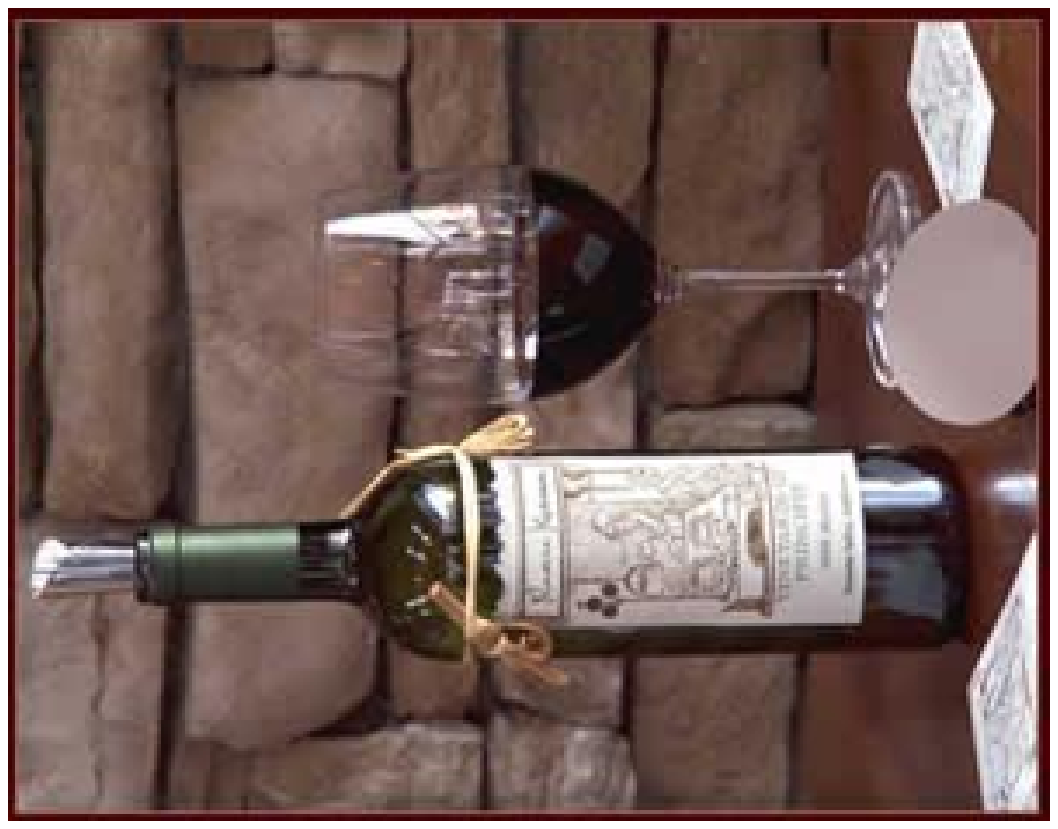}
\includegraphics[width=2.925cm,angle=90]{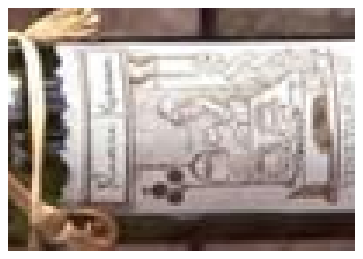}
\vspace*{-0.2cm}
\end{center}
\begin{center}
\includegraphics[height=2.925cm]{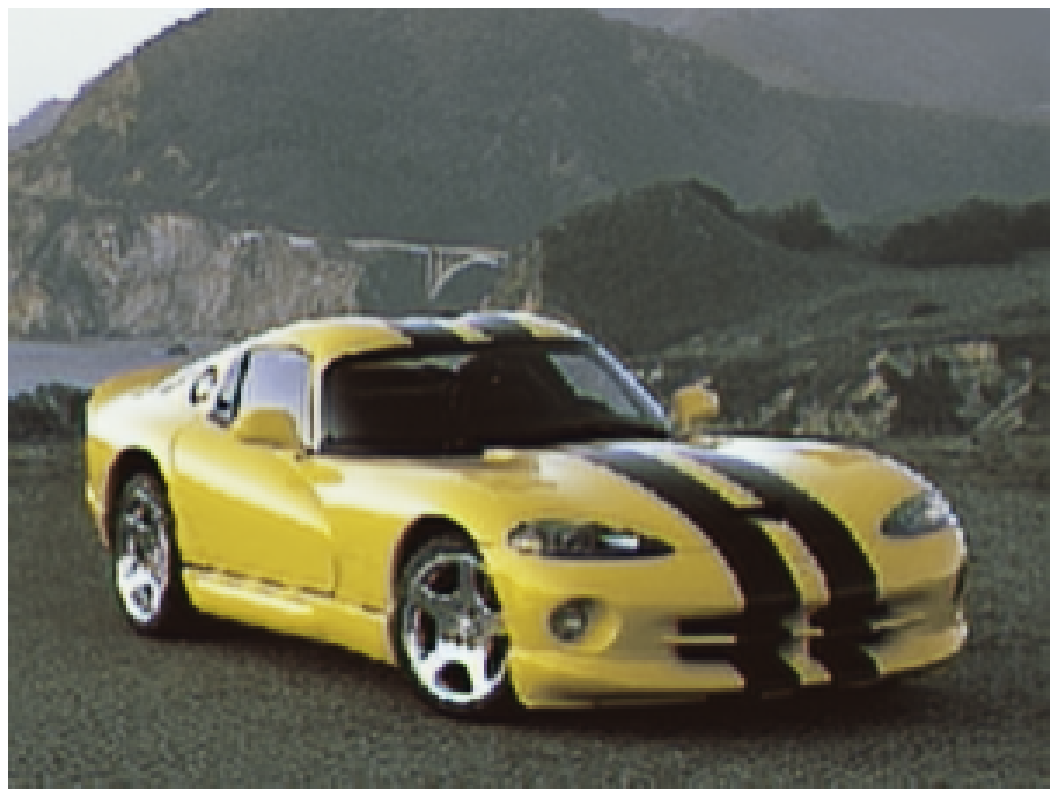}
\includegraphics[height=2.925cm]{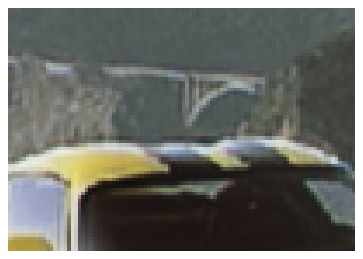}
\hspace*{0.3cm}
\includegraphics[width=2.925cm,angle=90]{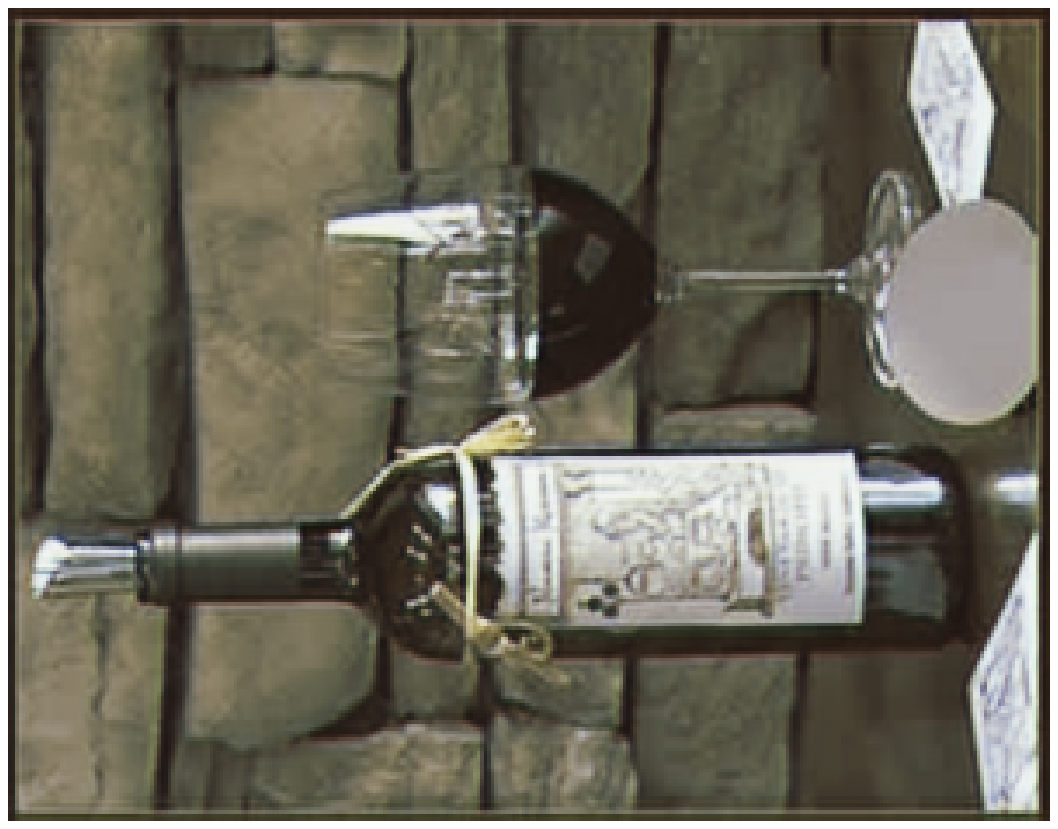}
\includegraphics[width=2.925cm,angle=90]{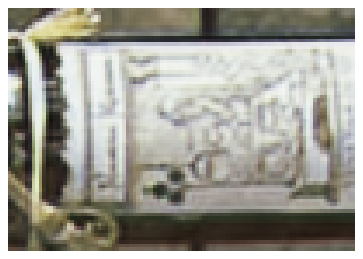}
\vspace*{-0.2cm}
\end{center}
\begin{center}
\includegraphics[height=2.925cm]{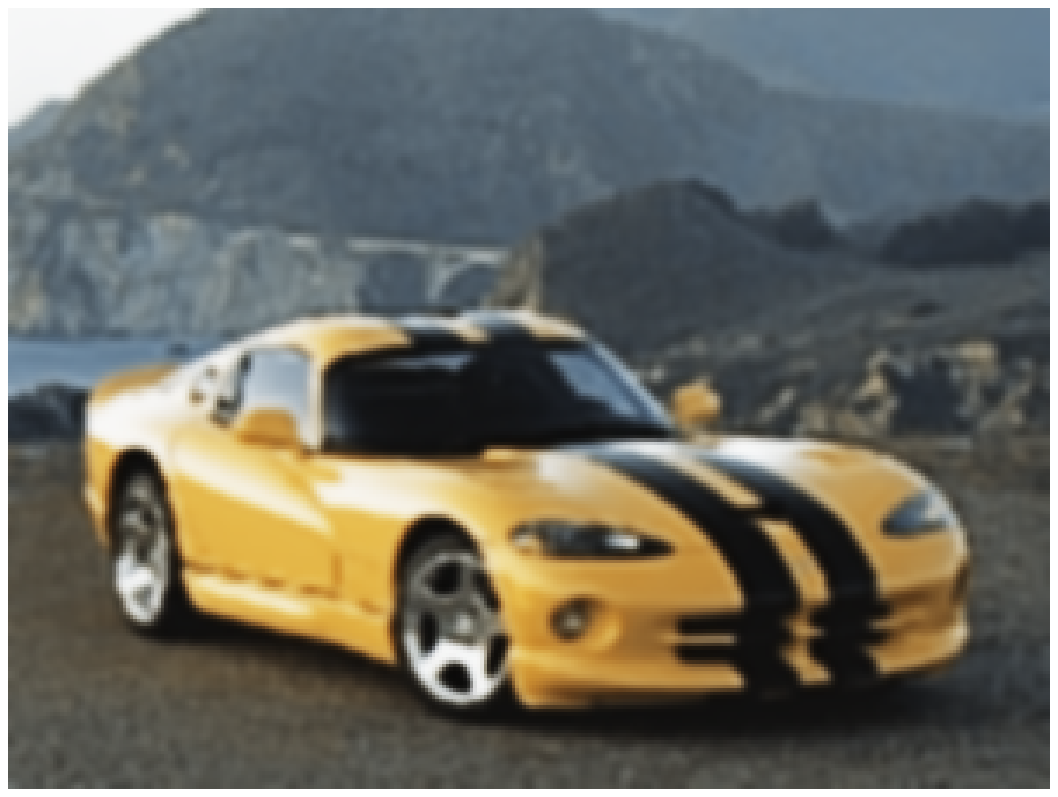}
\includegraphics[height=2.925cm]{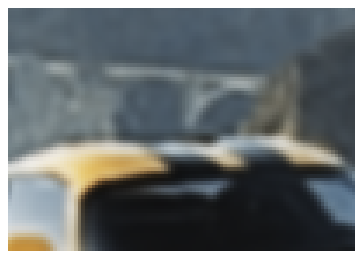}
\hspace*{0.3cm}
\includegraphics[width=2.925cm,angle=90]{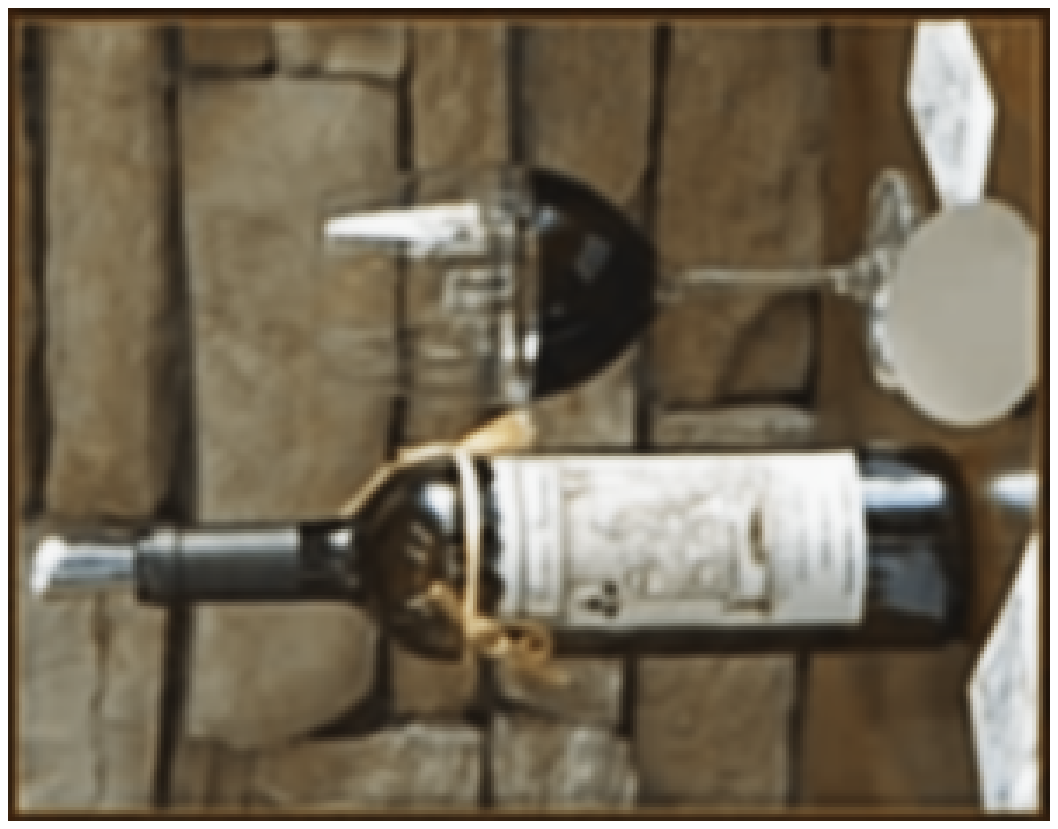}
\includegraphics[width=2.925cm,angle=90]{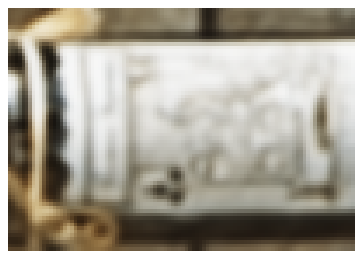}
\vspace*{-0.2cm}
\end{center}
\hspace*{4.1cm}(a)\hspace*{8.1cm}(b)
%
%
\begin{center}
\includegraphics[width=2.9cm,angle=90]{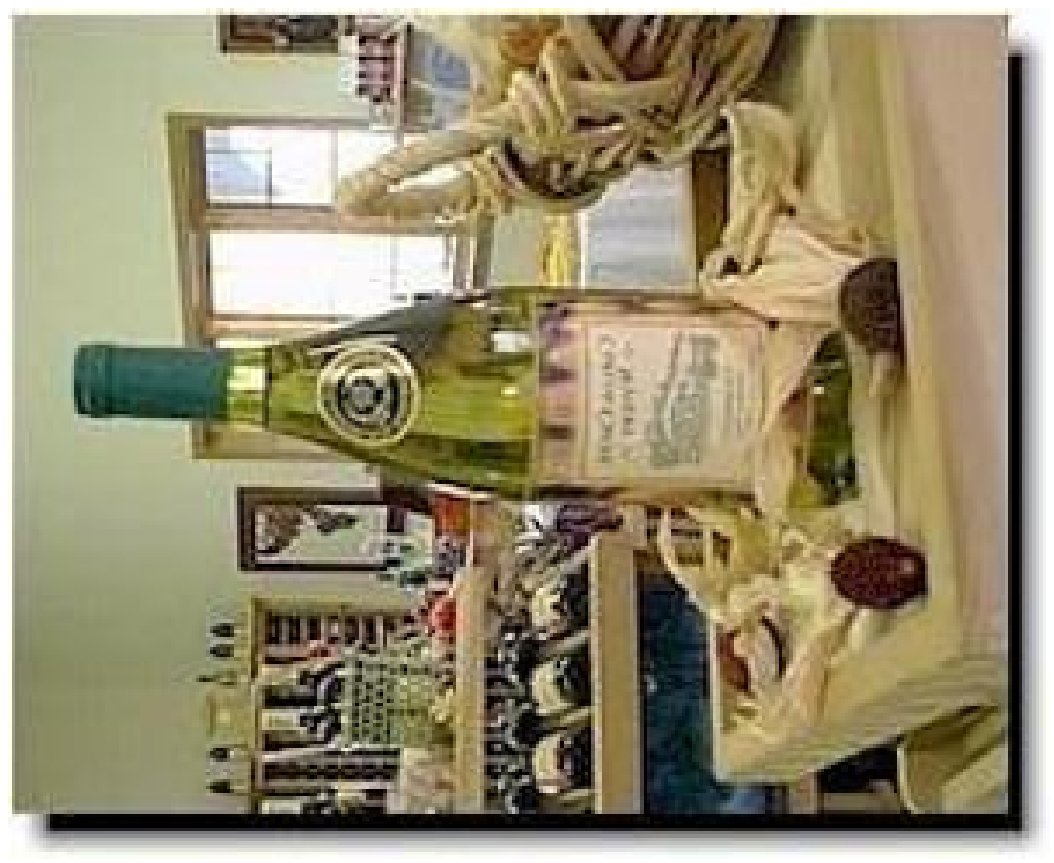}
\includegraphics[width=2.9cm,angle=90]{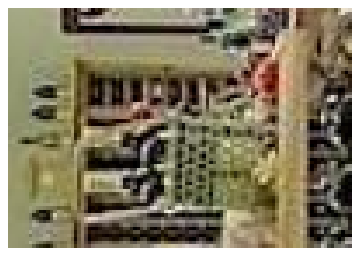}
\hspace*{0.3cm}
\includegraphics[height=2.9cm]{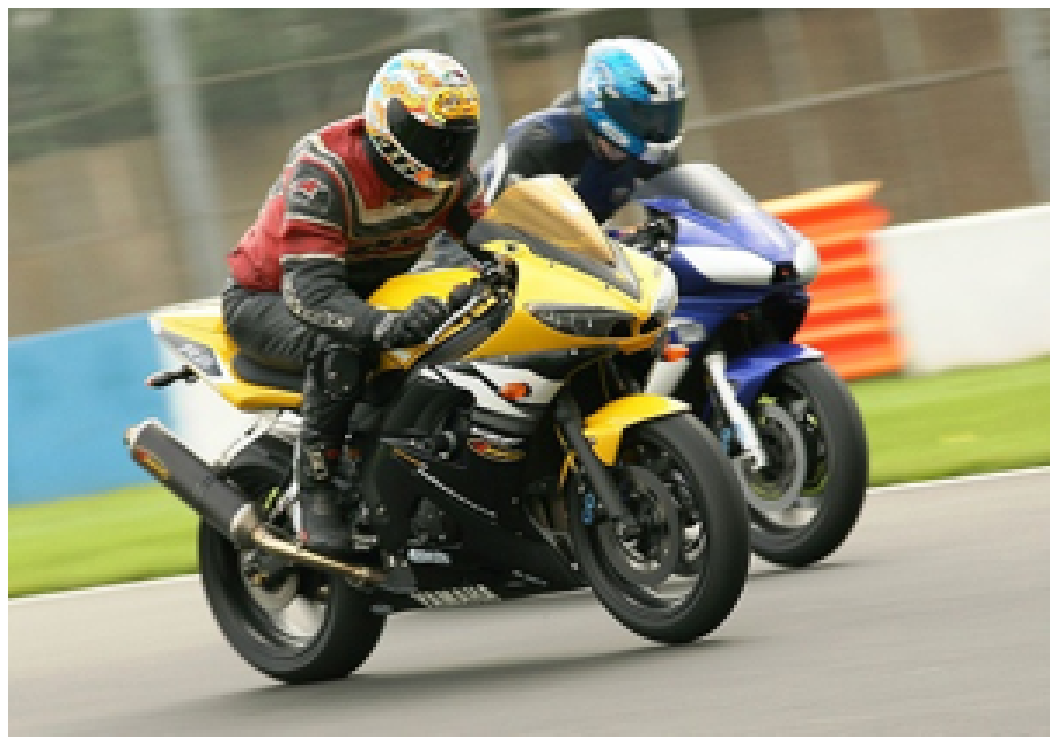}
\includegraphics[height=2.9cm]{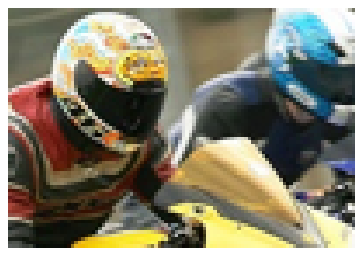}
\vspace*{-0.2cm}
\end{center}
\begin{center}
\includegraphics[width=2.9cm,angle=90]{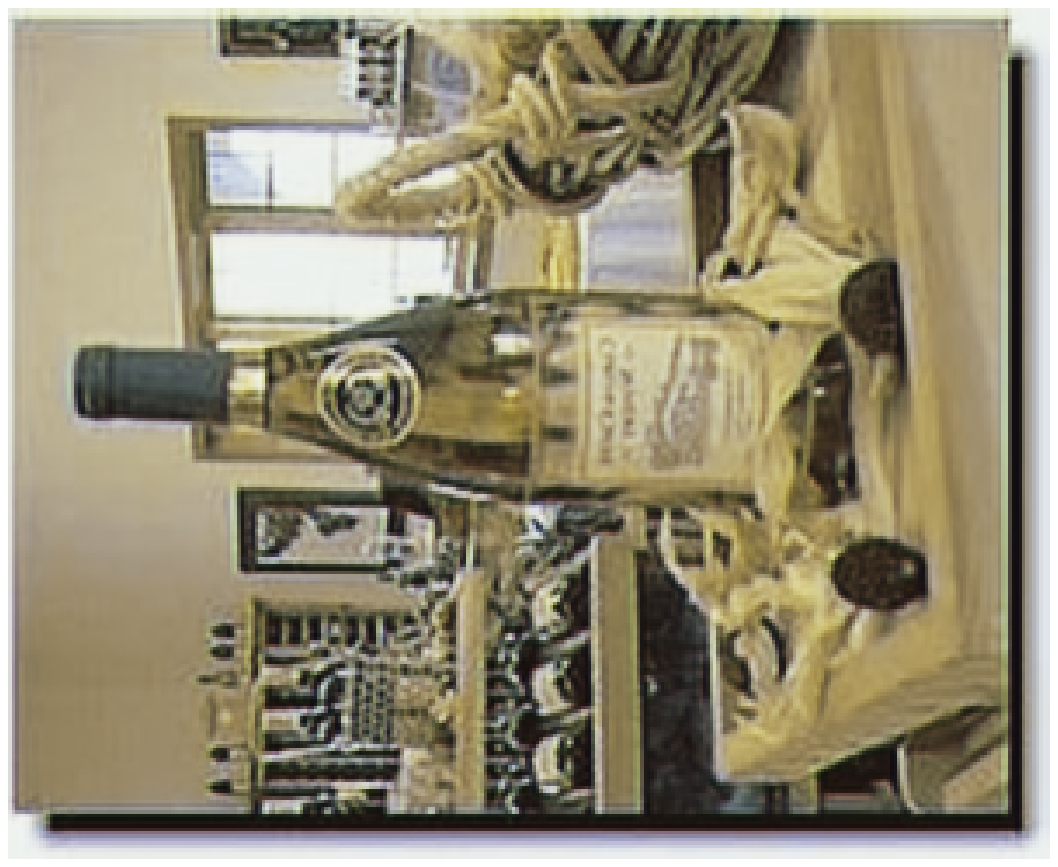}
\includegraphics[width=2.9cm,angle=90]{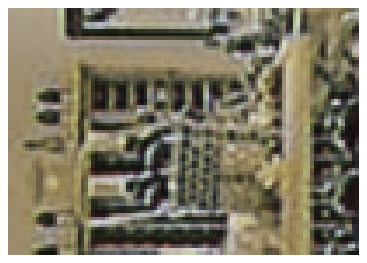}
\hspace*{0.3cm}
\includegraphics[height=2.9cm]{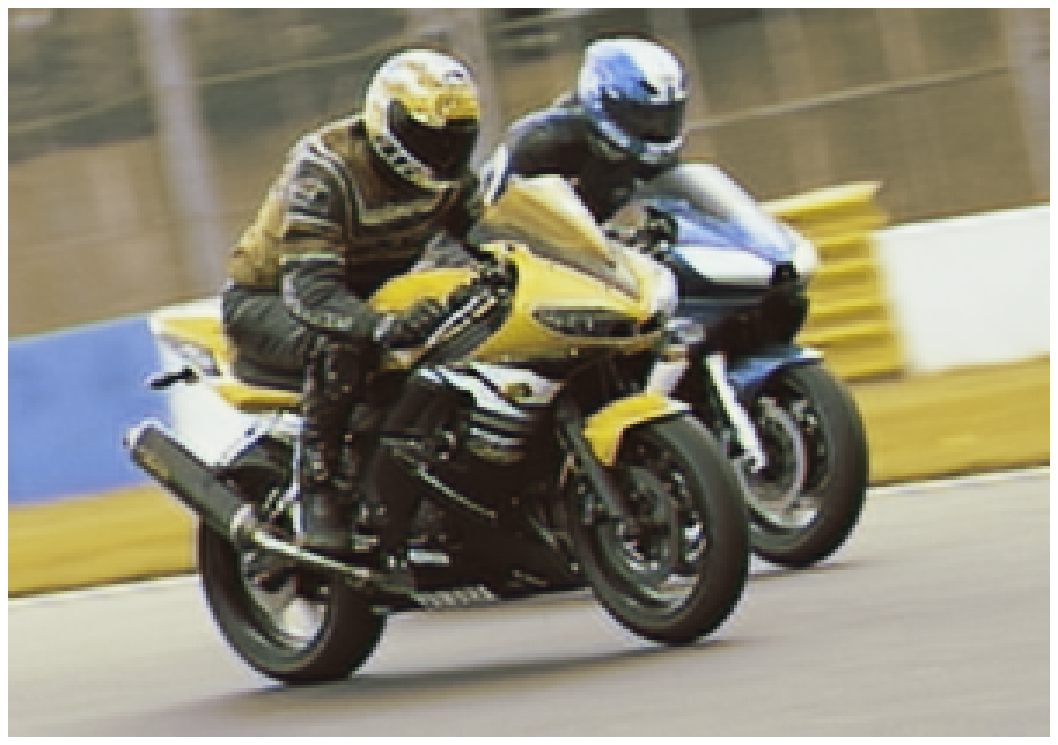}
\includegraphics[height=2.9cm]{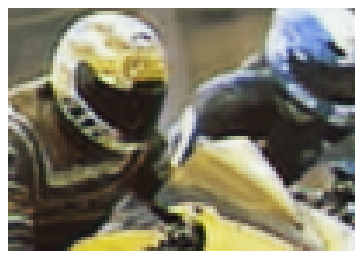}
\vspace*{-0.2cm}
\end{center}
\begin{center}
\includegraphics[width=2.9cm,angle=90]{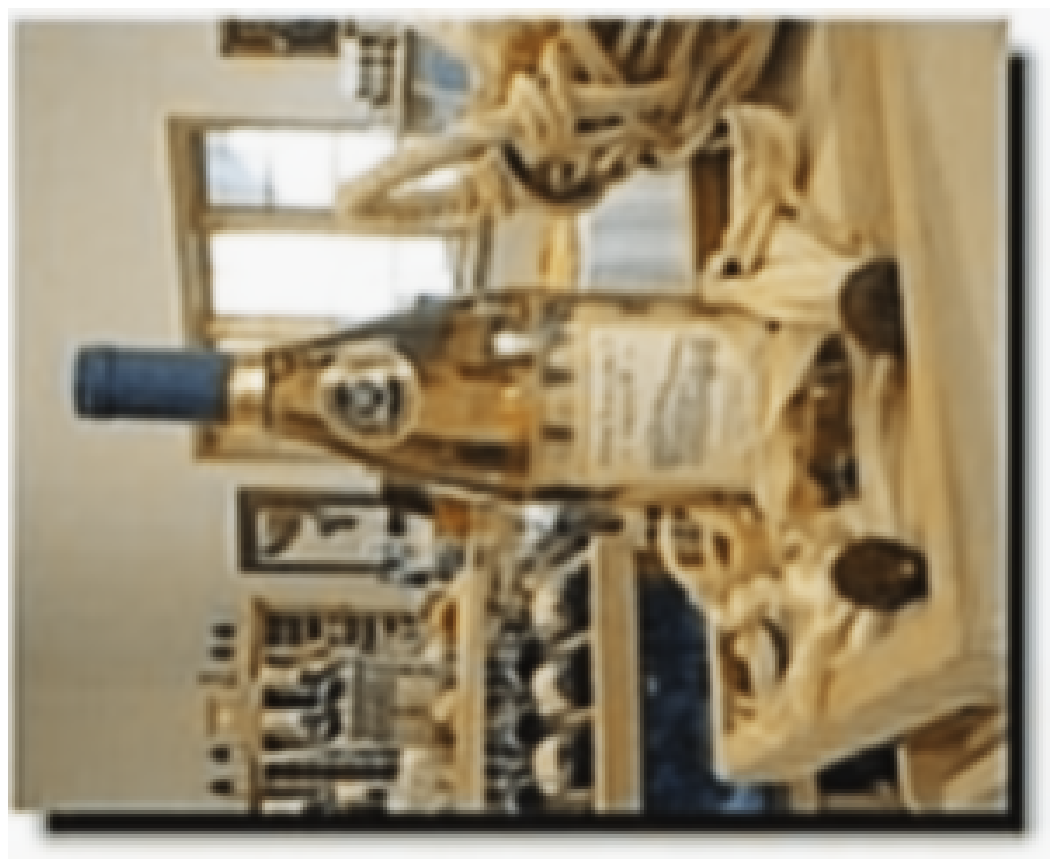}
\includegraphics[width=2.9cm,angle=90]{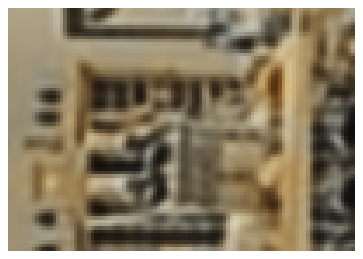}
\hspace*{0.3cm}
\includegraphics[height=2.9cm]{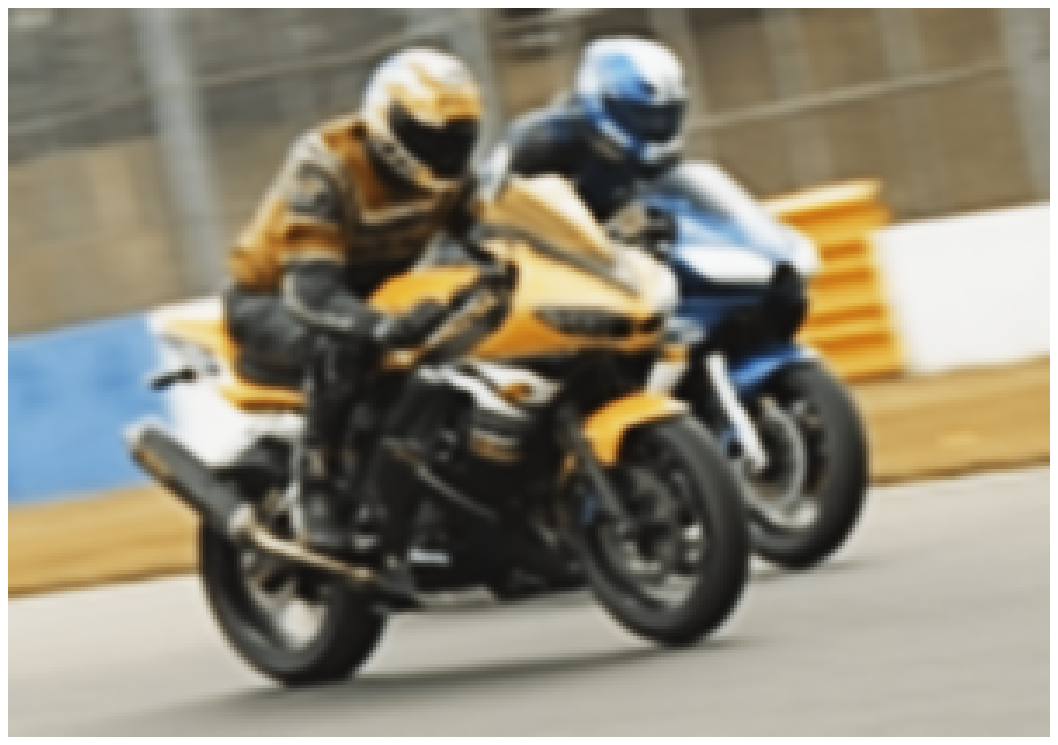}
\includegraphics[height=2.9cm]{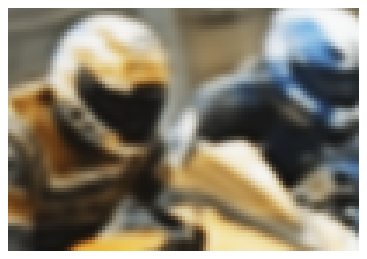}
\vspace*{-0.2cm}
\end{center}
\hspace*{4.1cm}(c)\hspace*{8.1cm}(d)
\vspace*{0.15cm}
\caption {
The original images and their reconstruction results. Top row: The
original images and close-up of their portion. Middle row: The
reconstructed images by the proposed method. Bottom row: The
reconstructed images by learning using $E_{PL}$. Introducing $E_{SFL}$
into learning is clearly effective to reduce the blurs.
}
\label{fig:reconstructed_images}
\vspace*{-0.2cm}
\hrulefill
\vspace*{-0.3cm}
\end{figure*}
%
%
%
\begin{figure*}[!t]
%
%
\begin{center}
\includegraphics[height=2.9cm]{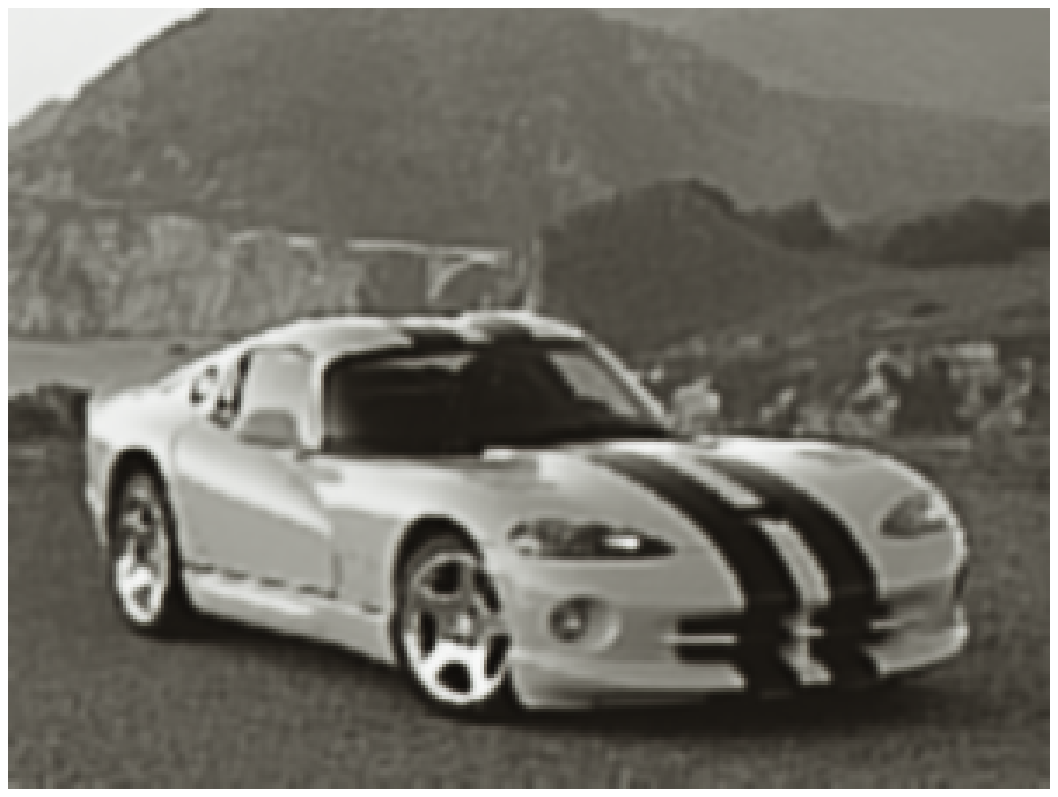}
\includegraphics[height=2.9cm]{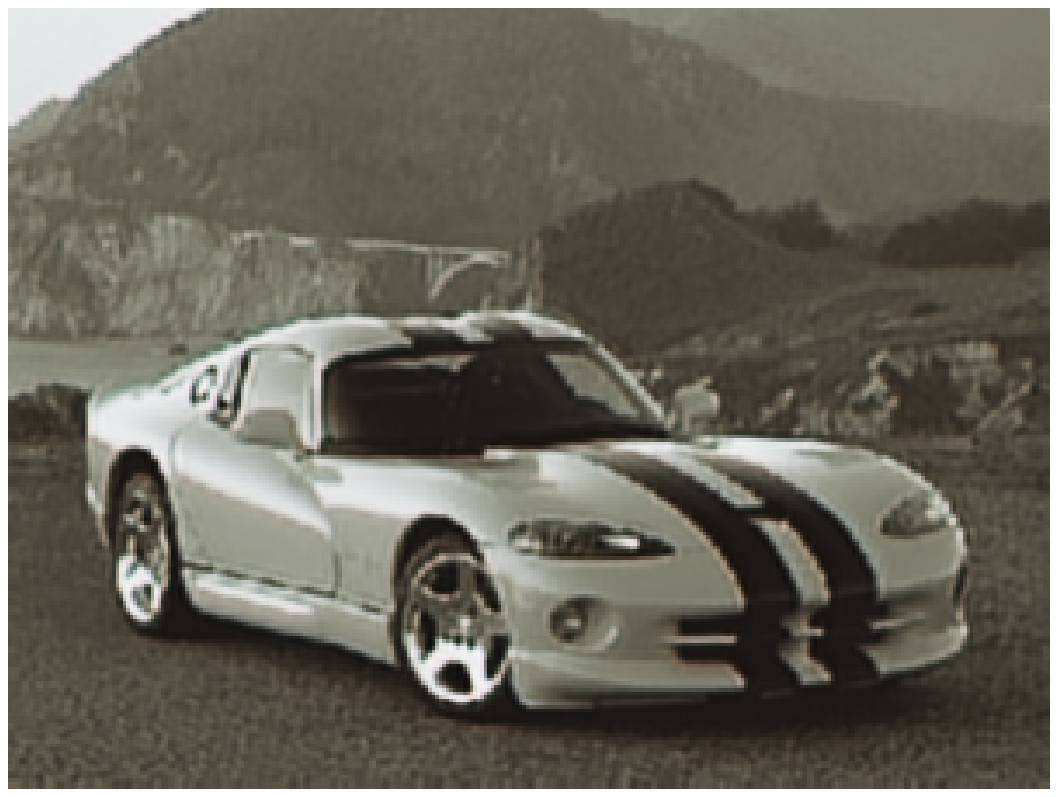}
\includegraphics[height=2.9cm]{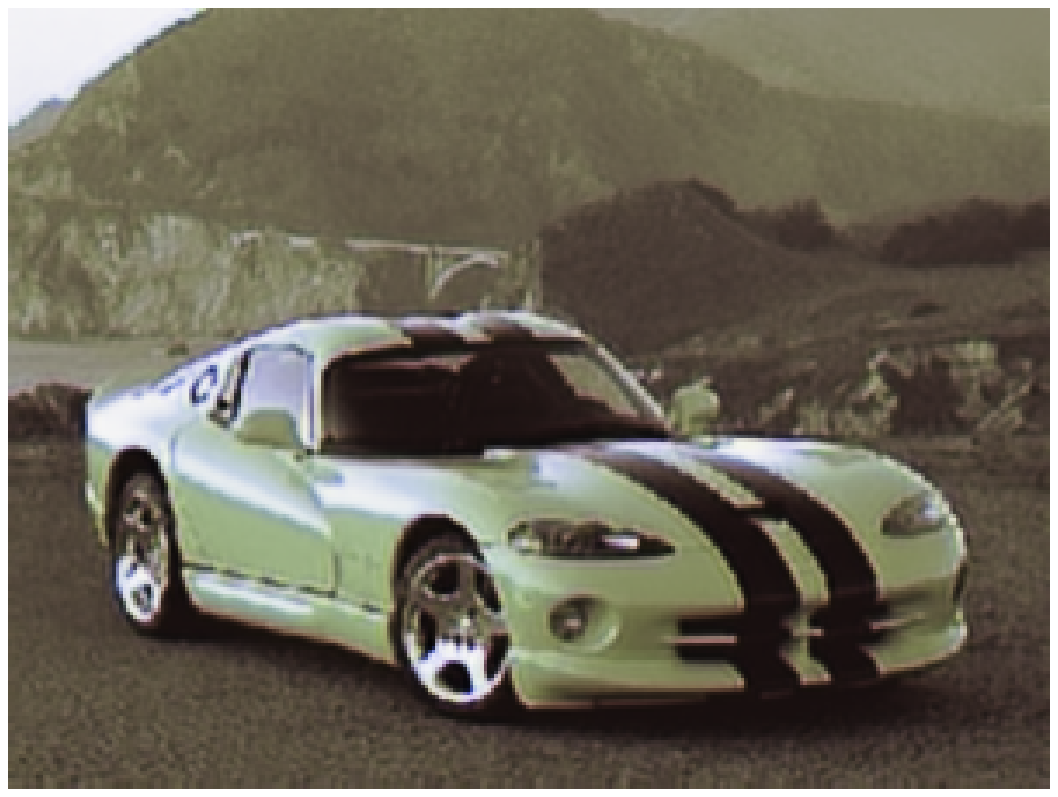}
\includegraphics[height=2.9cm]{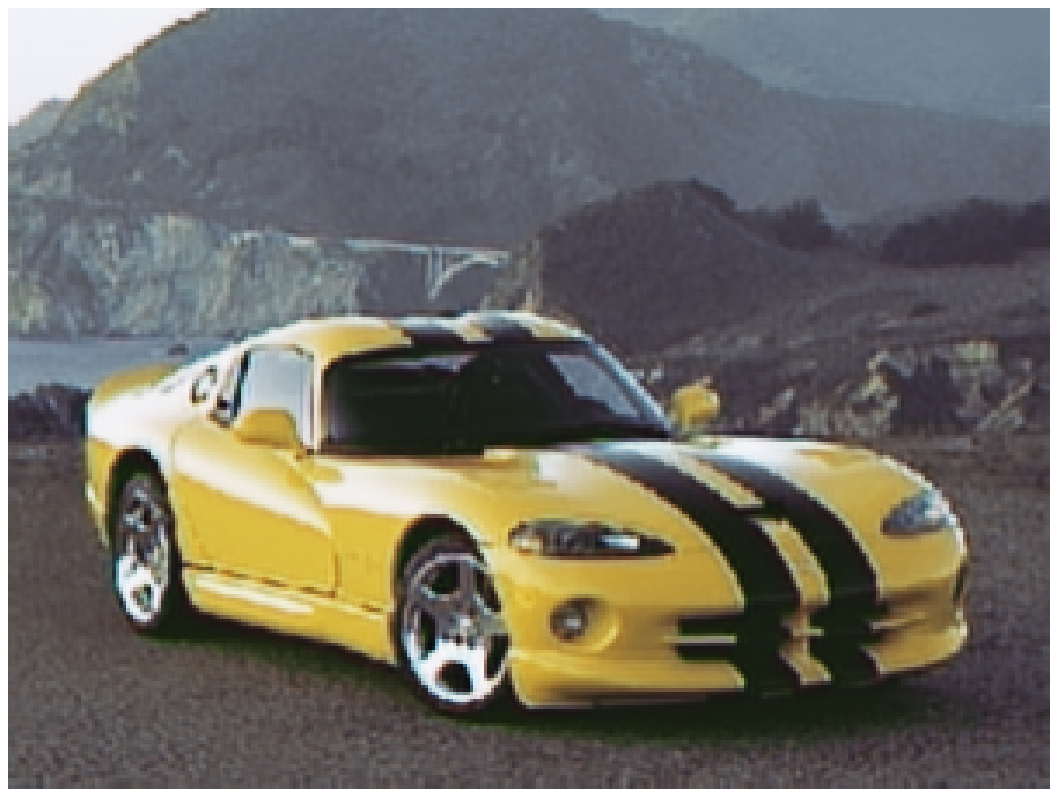}
\vspace*{-0.2cm}
\end{center}
\begin{center}
\includegraphics[height=2.9cm]{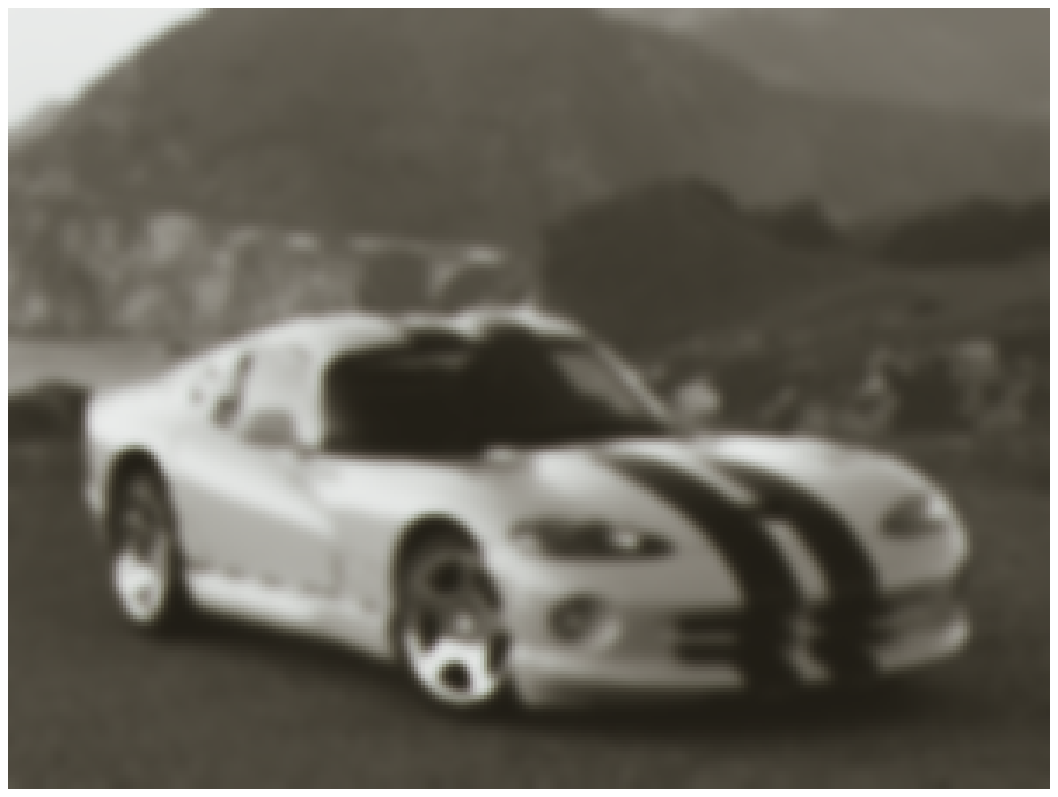}
\includegraphics[height=2.9cm]{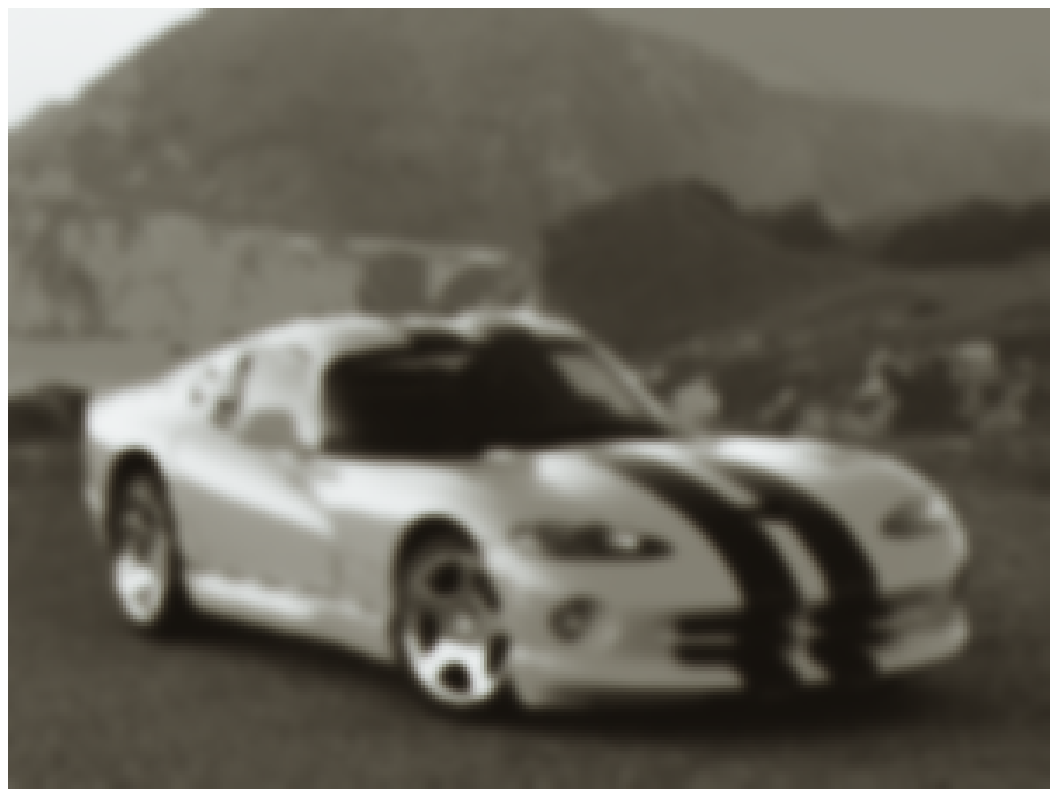}
\includegraphics[height=2.9cm]{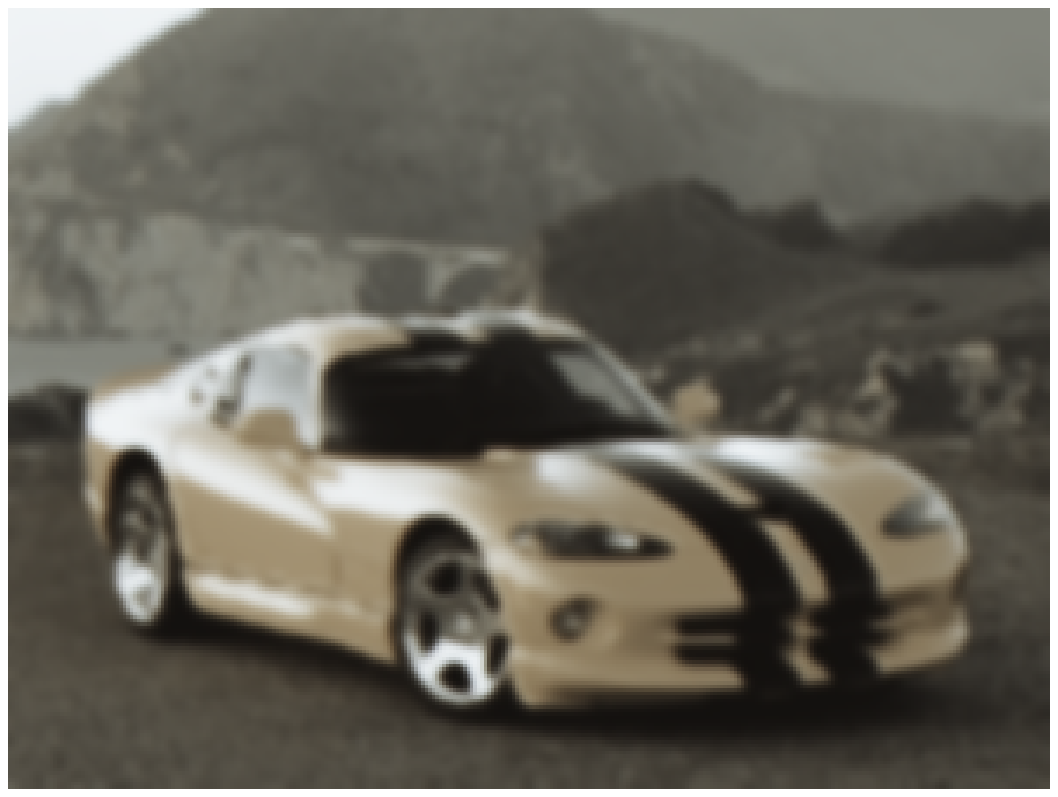}
\includegraphics[height=2.9cm]{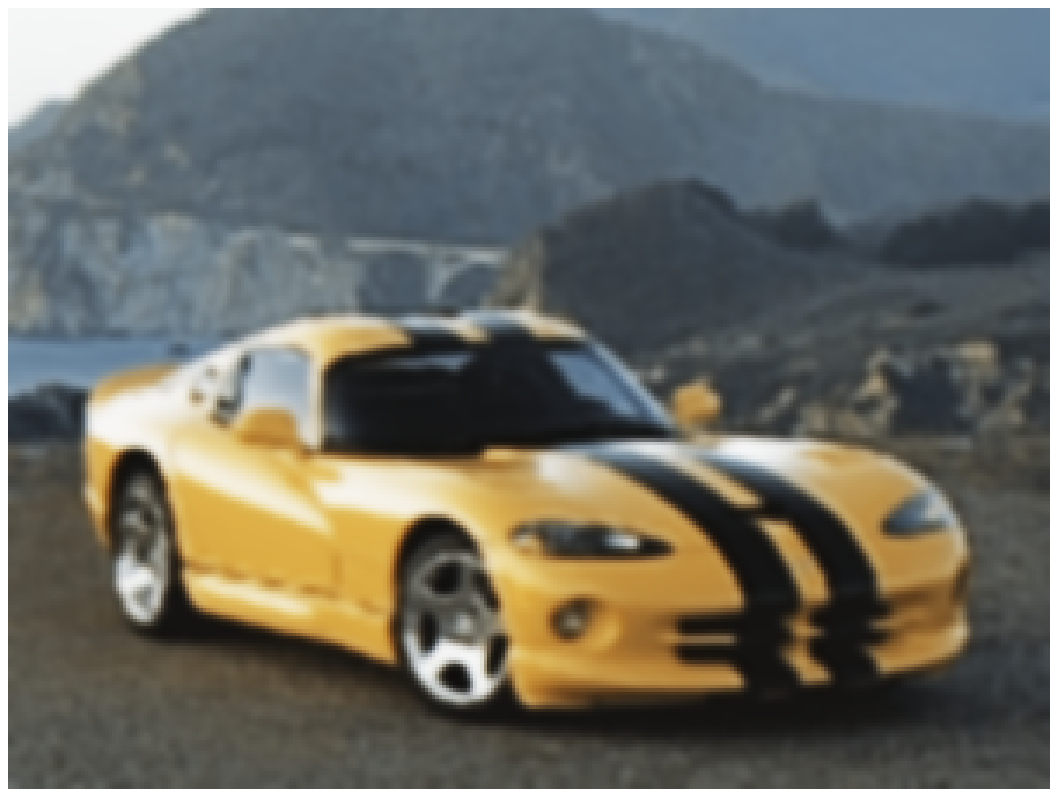}
\vspace*{-0.2cm}
\end{center}
\hspace*{8.45cm}(a)
%
%
\begin{center}
\includegraphics[width=3.02cm,angle=90]{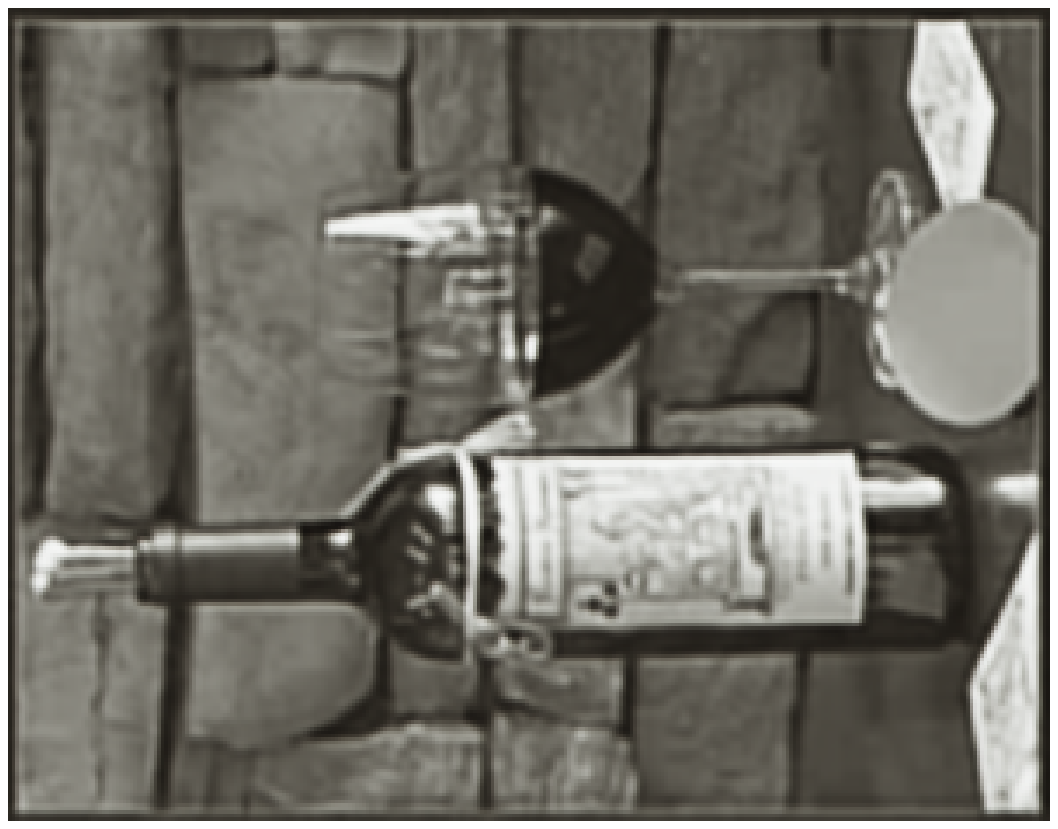}
\includegraphics[width=3.02cm,angle=90]{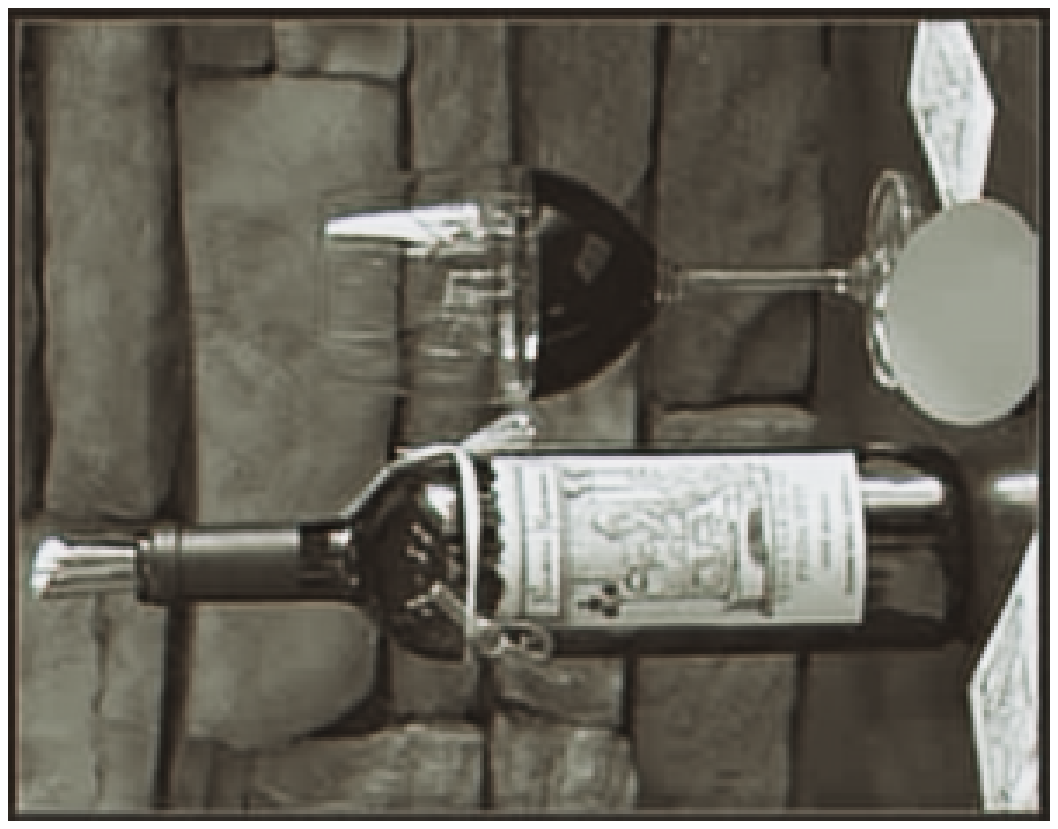}
\includegraphics[width=3.02cm,angle=90]{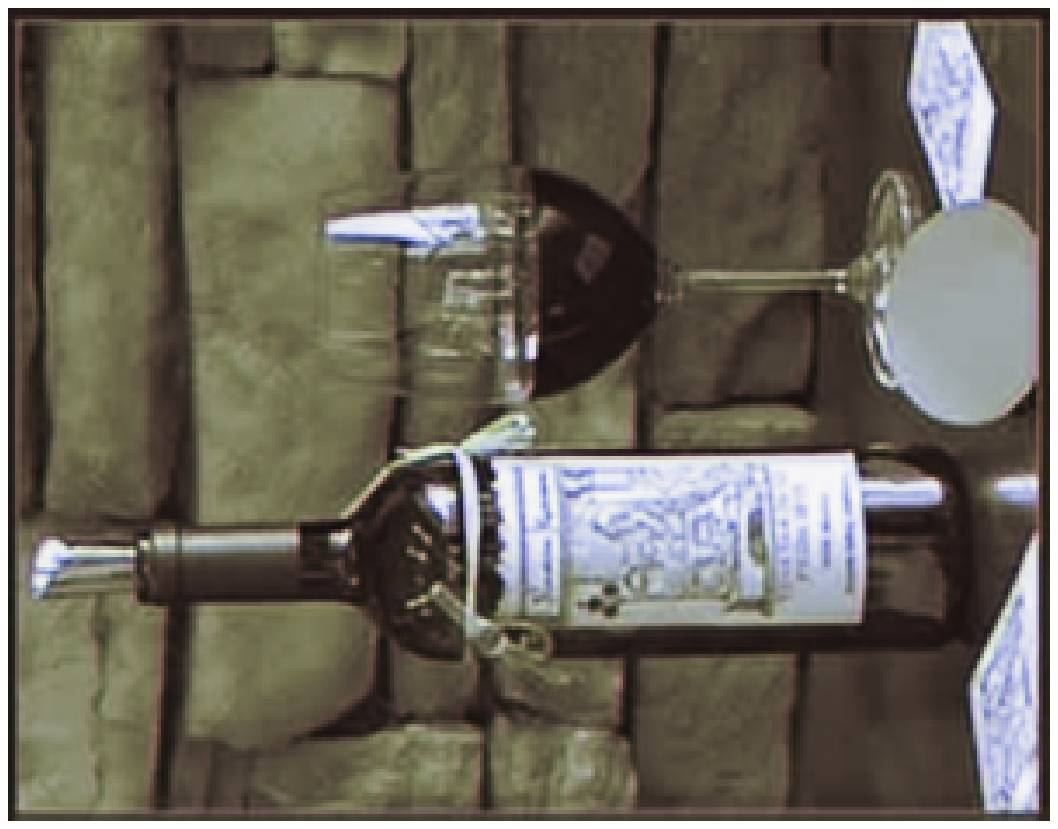}
\includegraphics[width=3.02cm,angle=90]{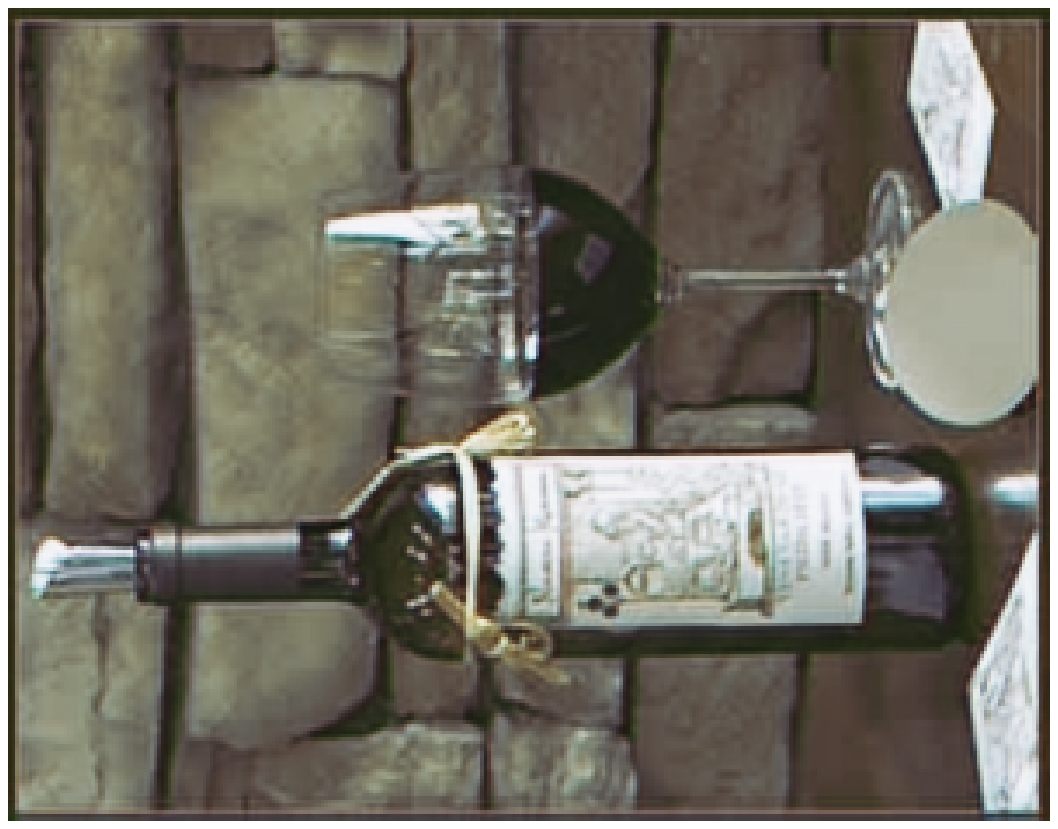}
\vspace*{-0.2cm}
\end{center}
\begin{center}
\includegraphics[width=3.02cm,angle=90]{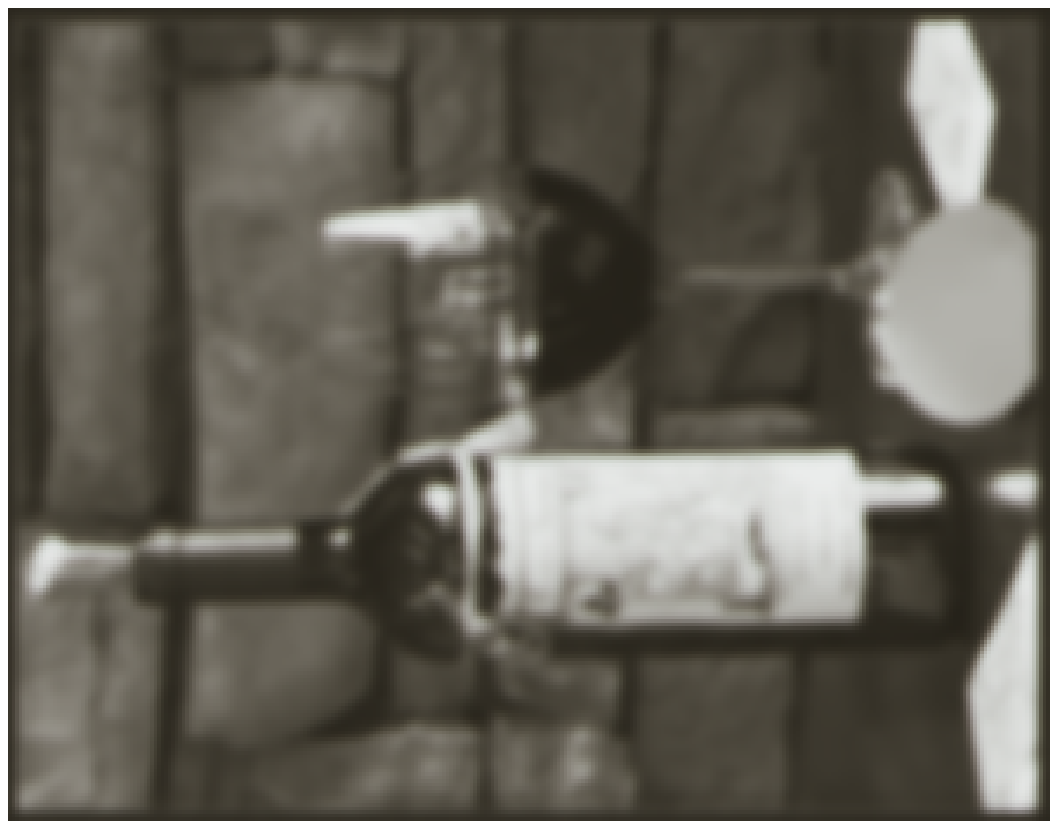}
\includegraphics[width=3.02cm,angle=90]{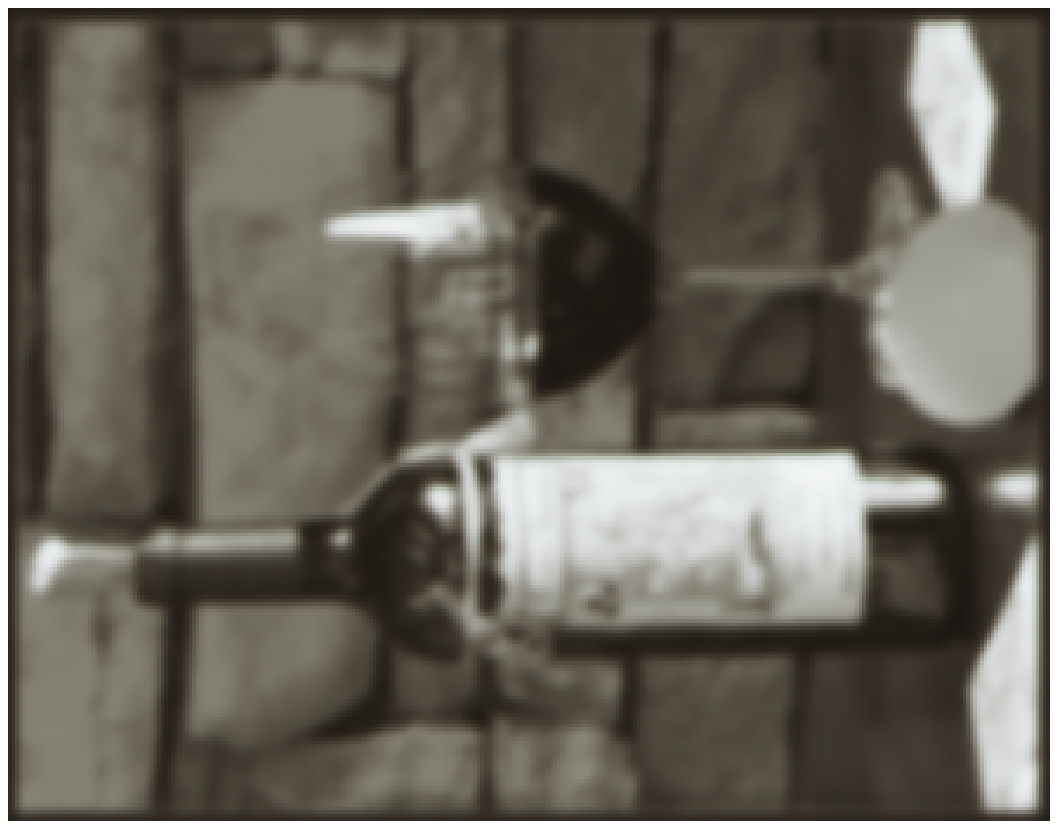}
\includegraphics[width=3.02cm,angle=90]{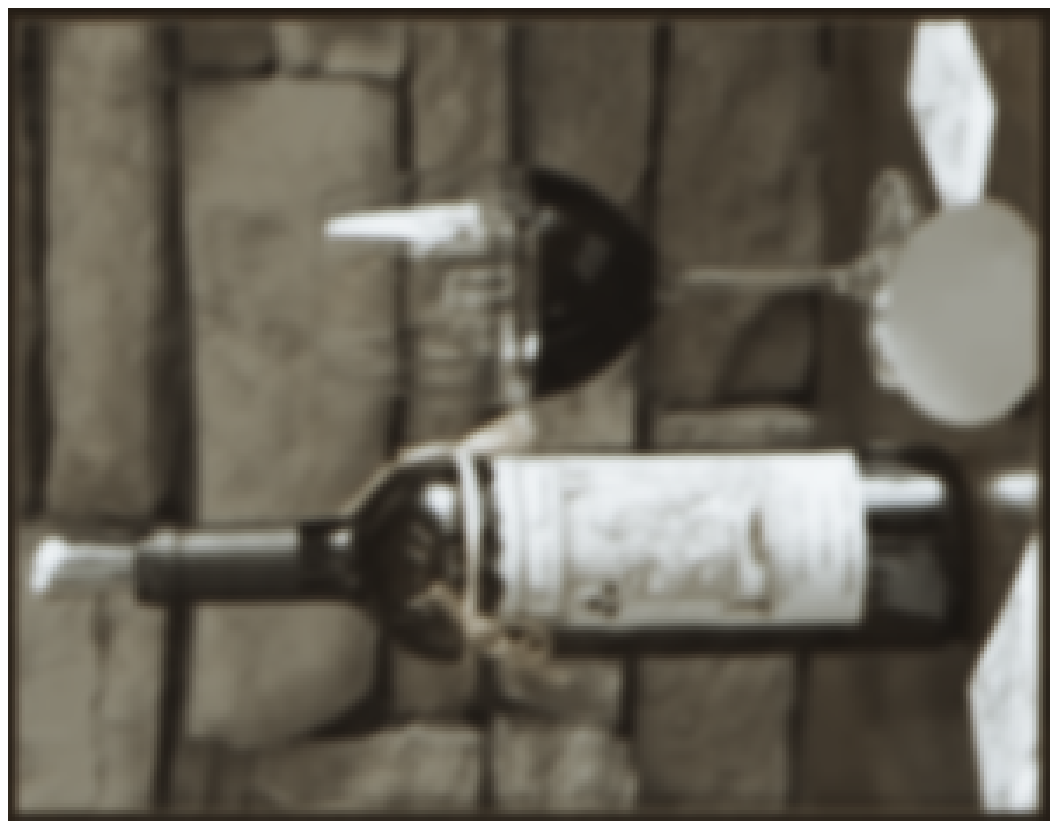}
\includegraphics[width=3.02cm,angle=90]{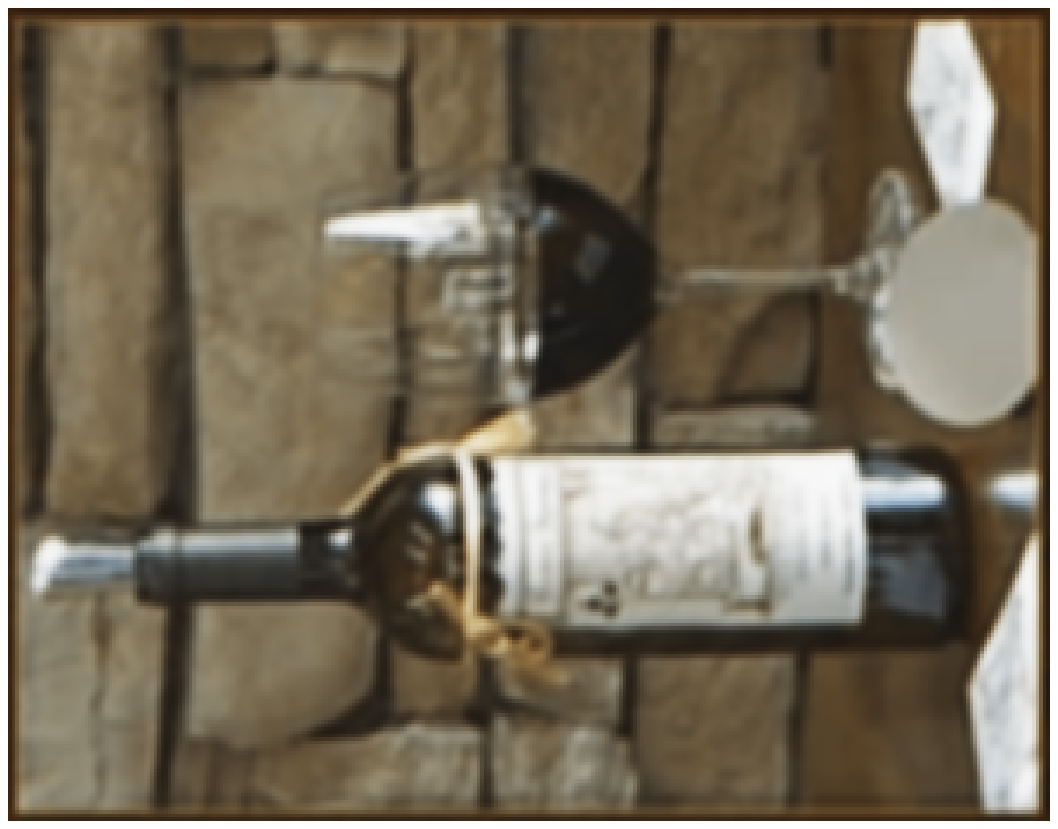}
\vspace*{-0.2cm}
\end{center}
\hspace*{8.45cm}(b)
\vspace*{0.2cm}
\caption {
The reconstructed images for the images in
Fig.\ref{fig:reconstructed_images}(a) and (b) at 100, 500, 1000 and
1500 epochs. Top row: The reconstructed images by the proposed
method. Bottom row: The reconstructed images by learning using
$E_{PL}$.
}
%
%
\vspace*{-0.2cm}
\hrulefill
\vspace*{-0.3cm}
\end{figure*}
%
%
%
\addtocounter{figure}{-1}
\begin{figure*}[!t]
%
%
\begin{center}
\includegraphics[width=3.4cm,angle=90]{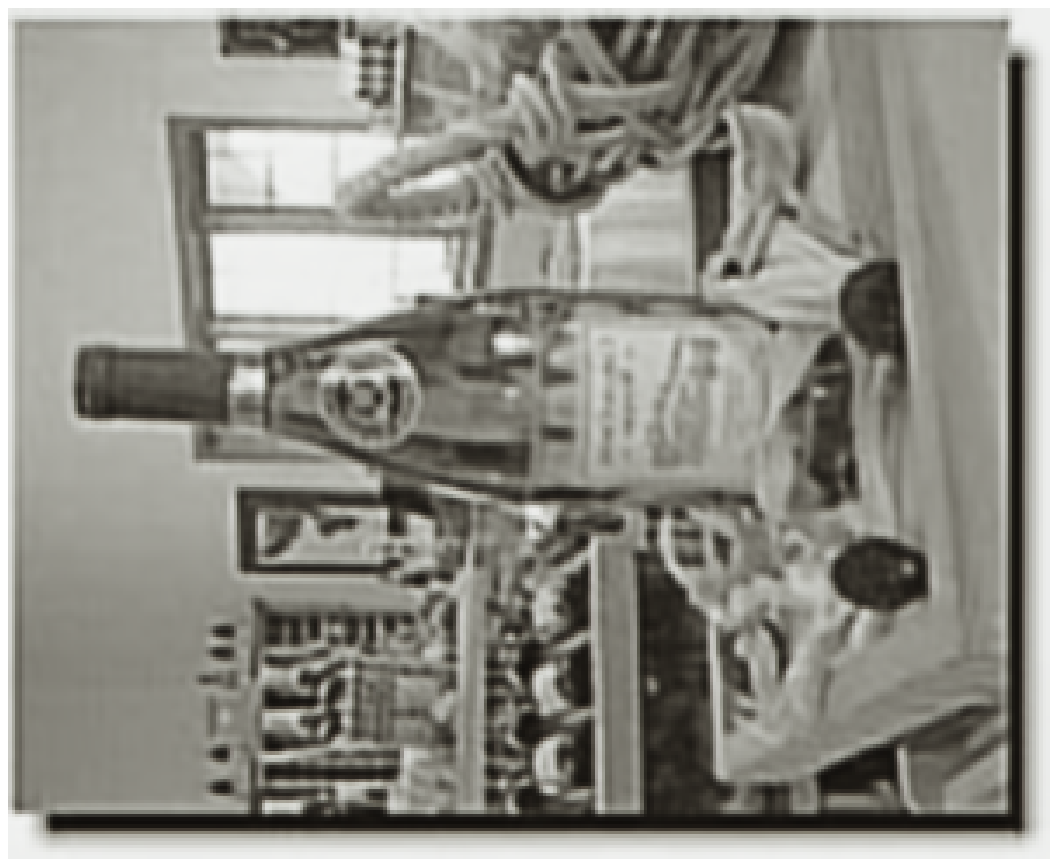}
\includegraphics[width=3.4cm,angle=90]{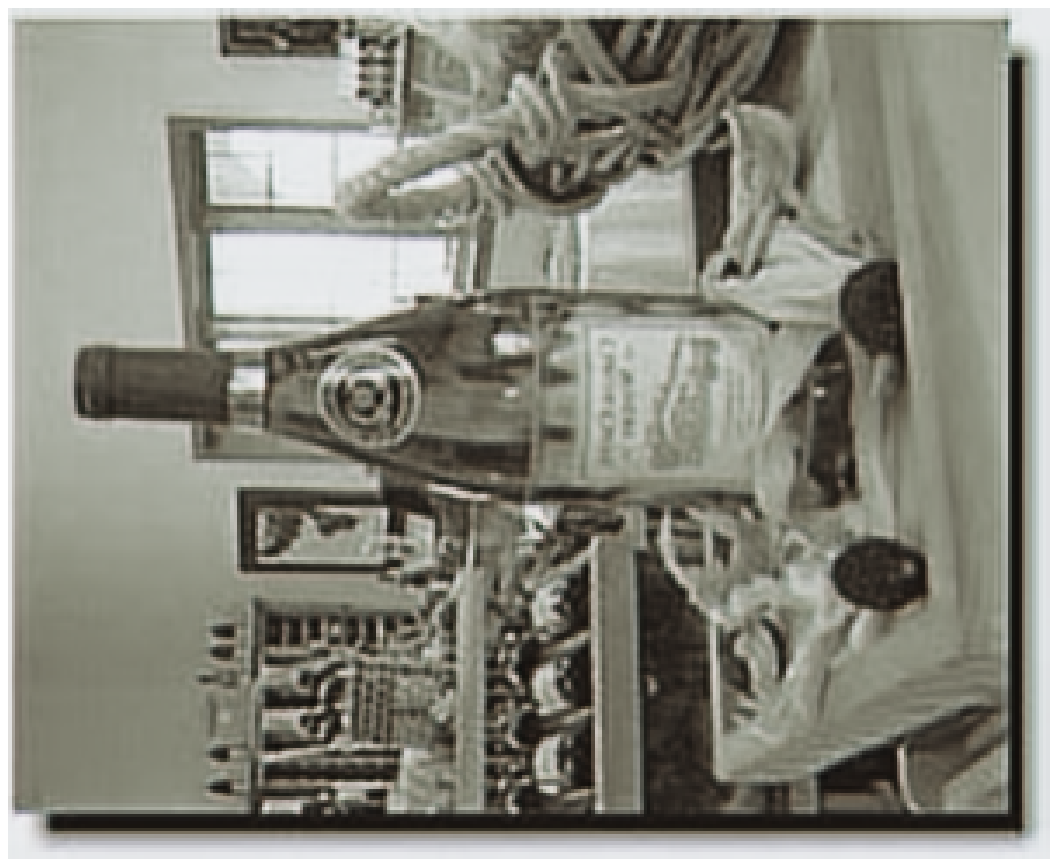}
\includegraphics[width=3.4cm,angle=90]{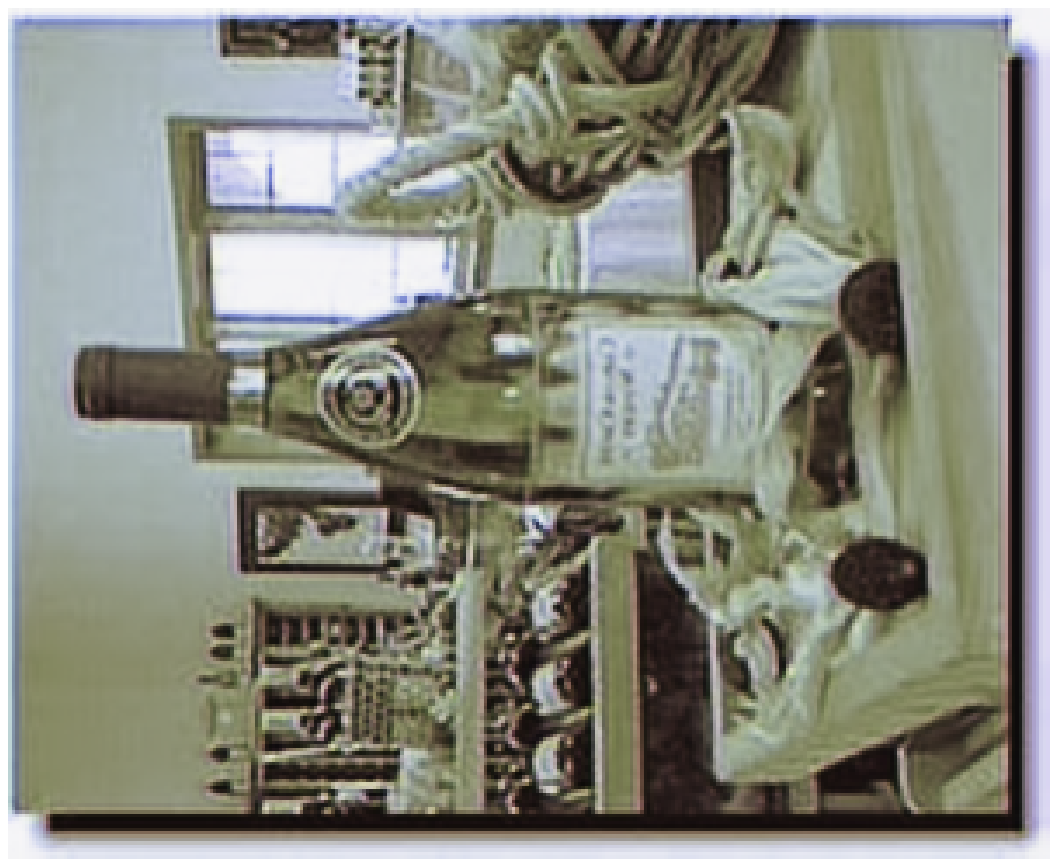}
\includegraphics[width=3.4cm,angle=90]{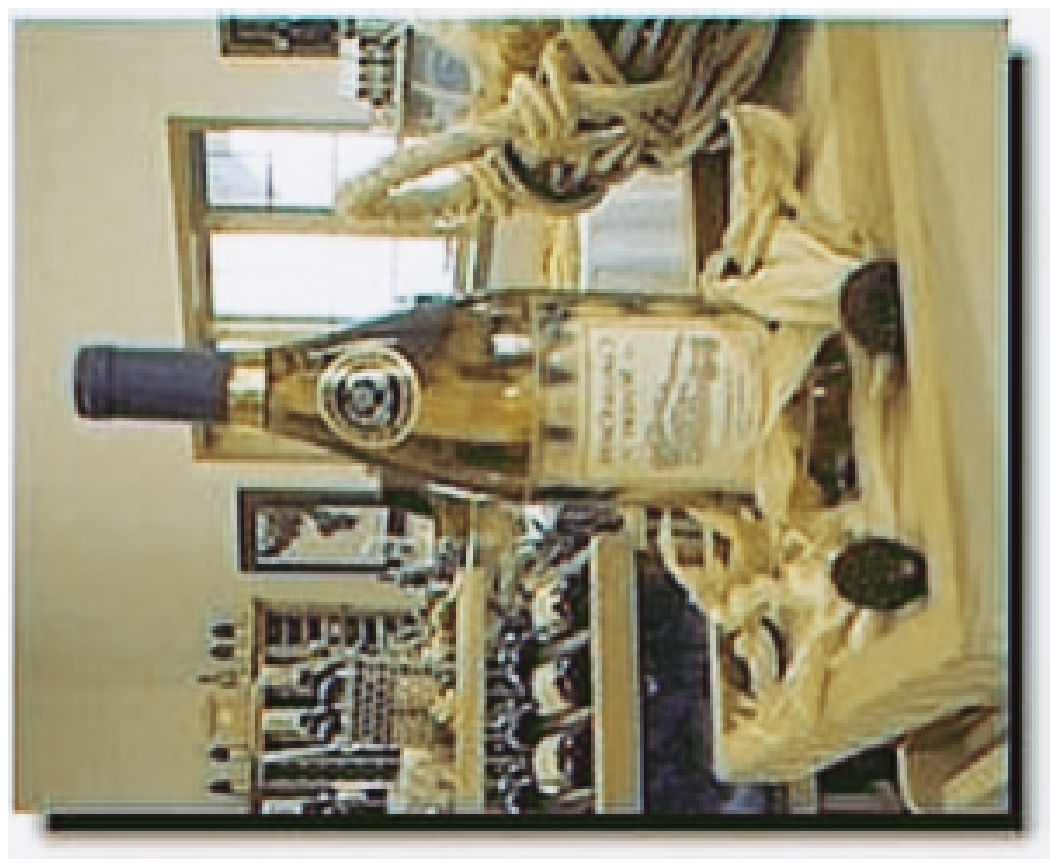}
\vspace*{-0.2cm}
\end{center}
\begin{center}
\includegraphics[width=3.4cm,angle=90]{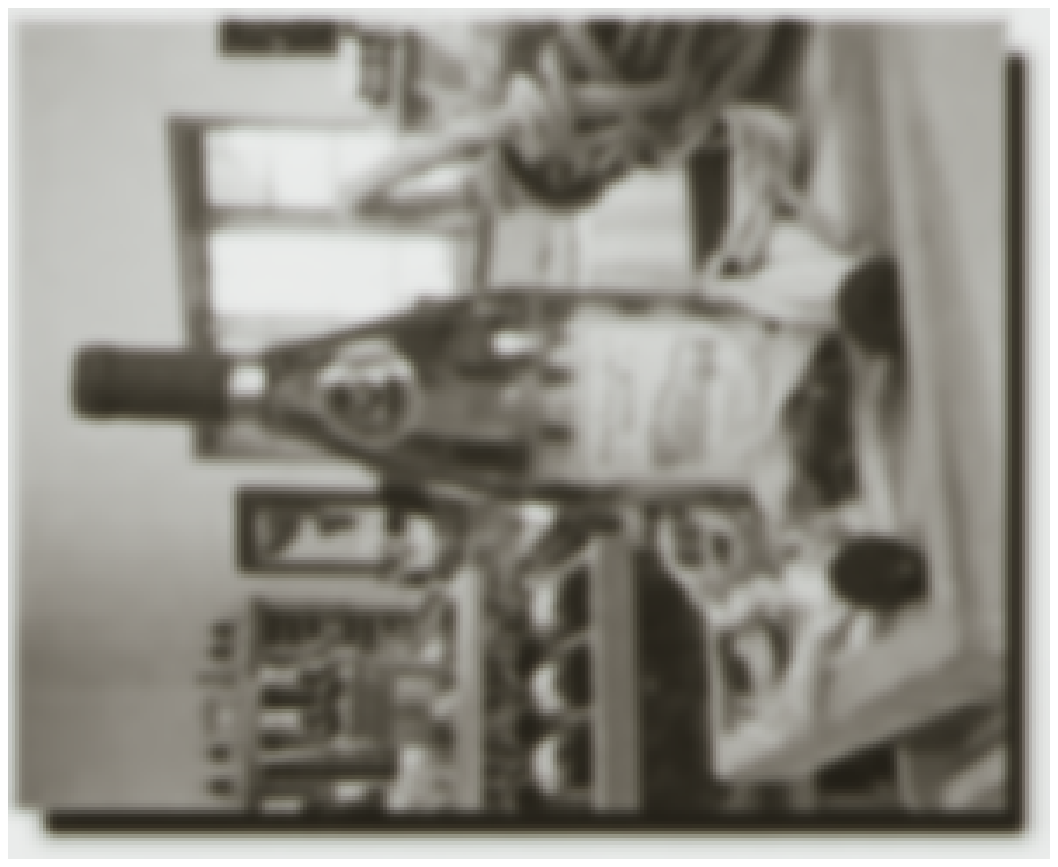}
\includegraphics[width=3.4cm,angle=90]{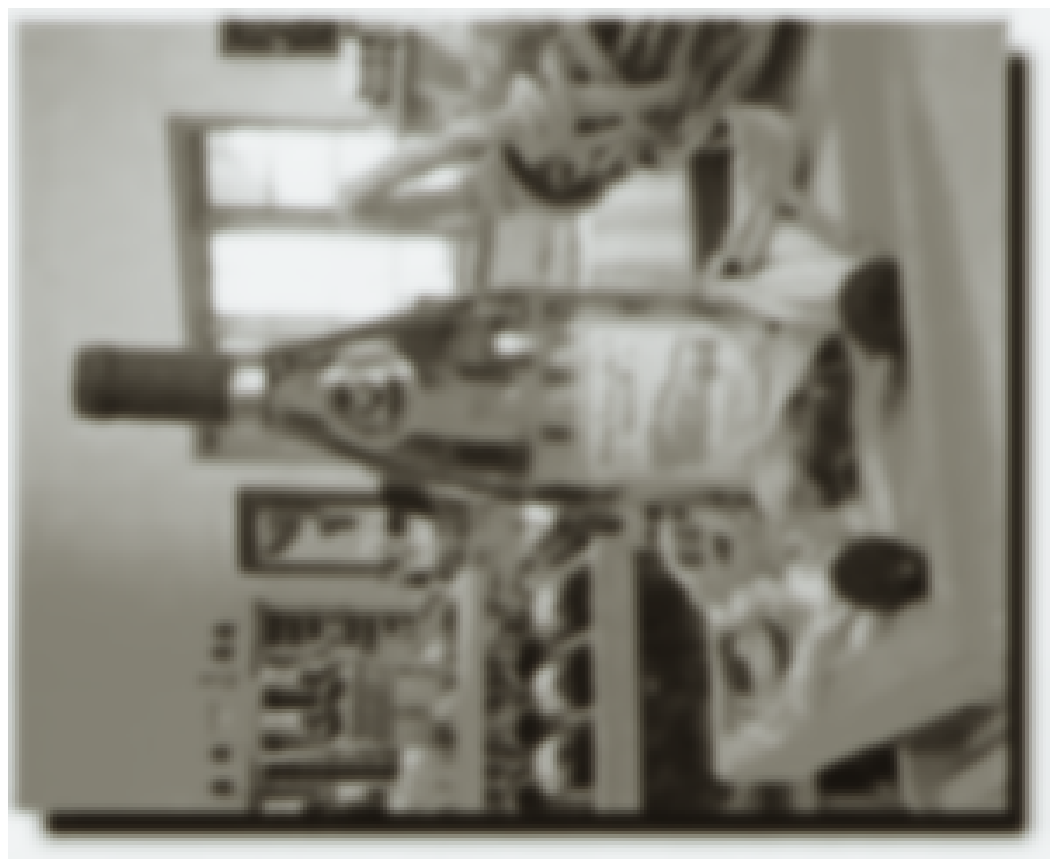}
\includegraphics[width=3.4cm,angle=90]{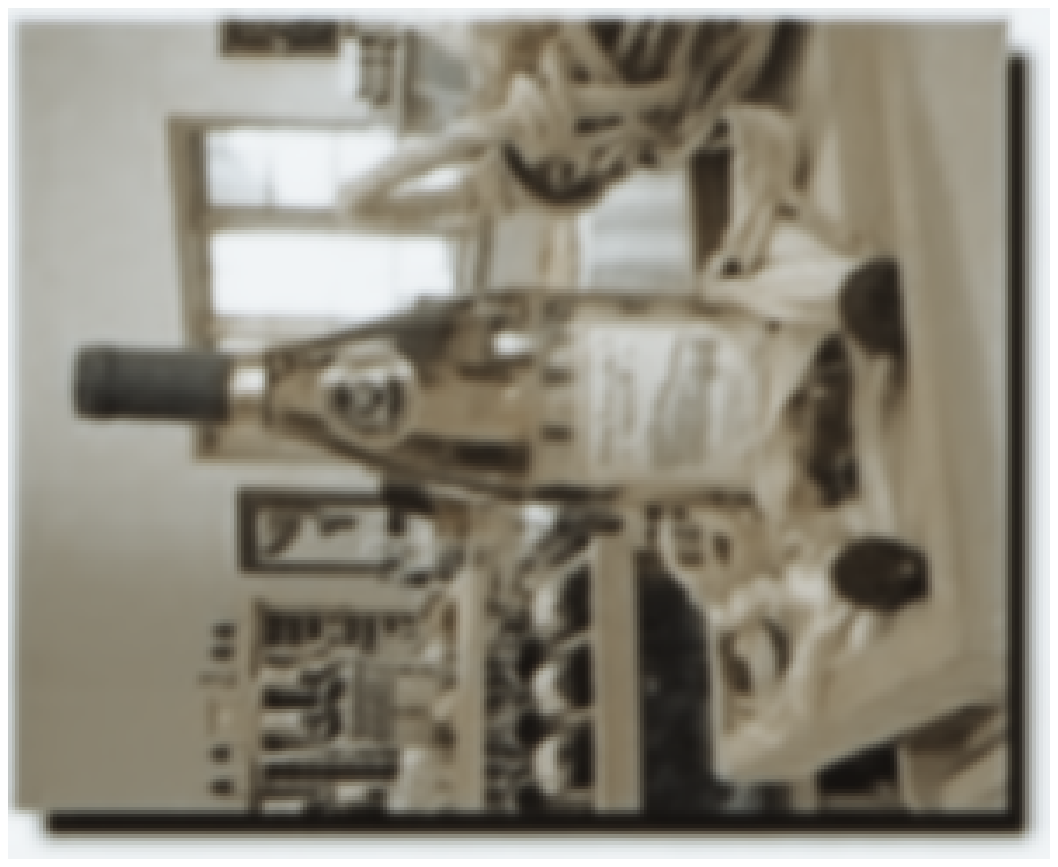}
\includegraphics[width=3.4cm,angle=90]{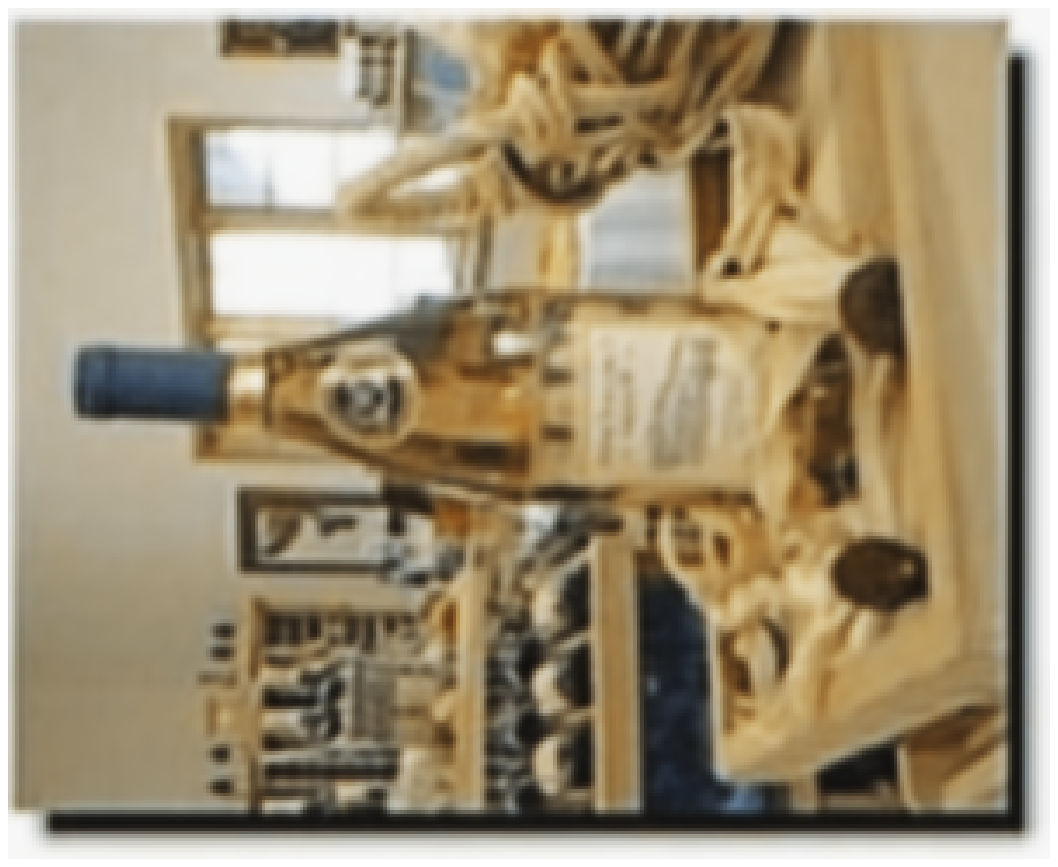}
\vspace*{-0.2cm}
\end{center}
\hspace*{8.45cm}(c)
%
%
\begin{center}
\includegraphics[height=2.9cm]{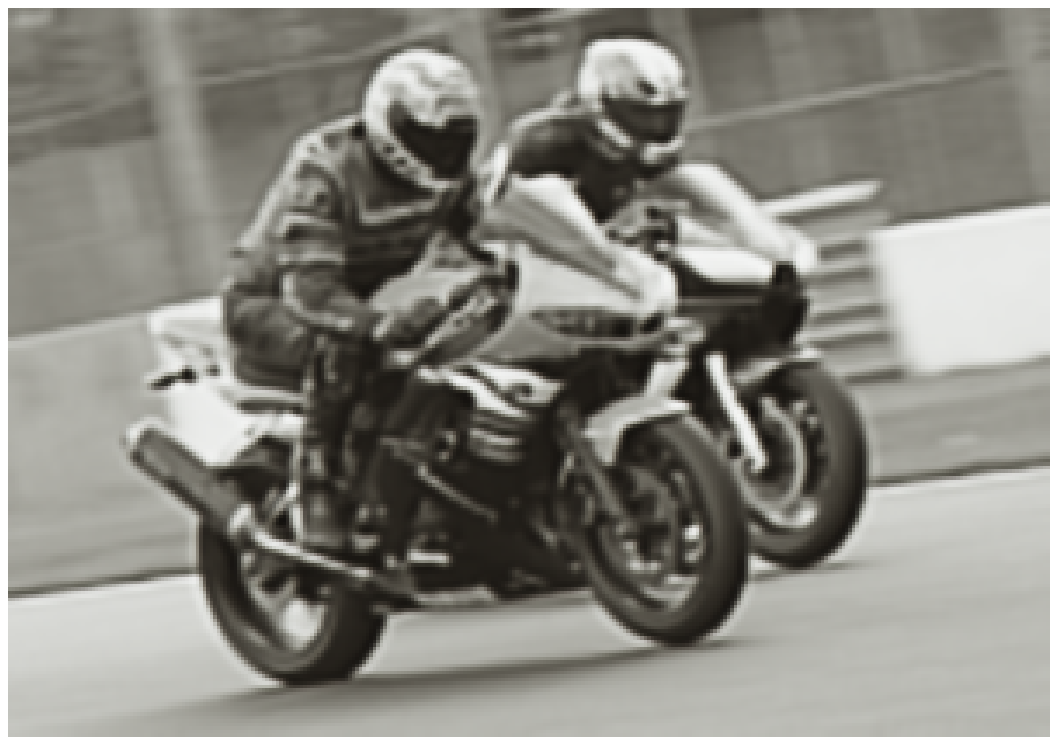}
\includegraphics[height=2.9cm]{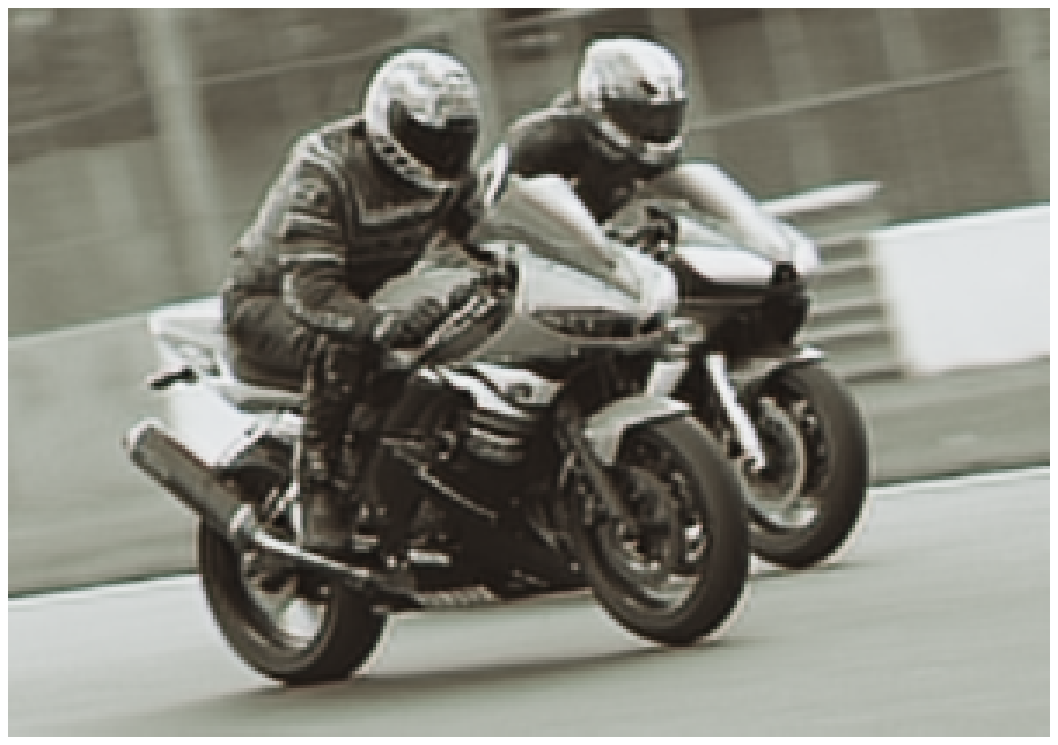}
\includegraphics[height=2.9cm]{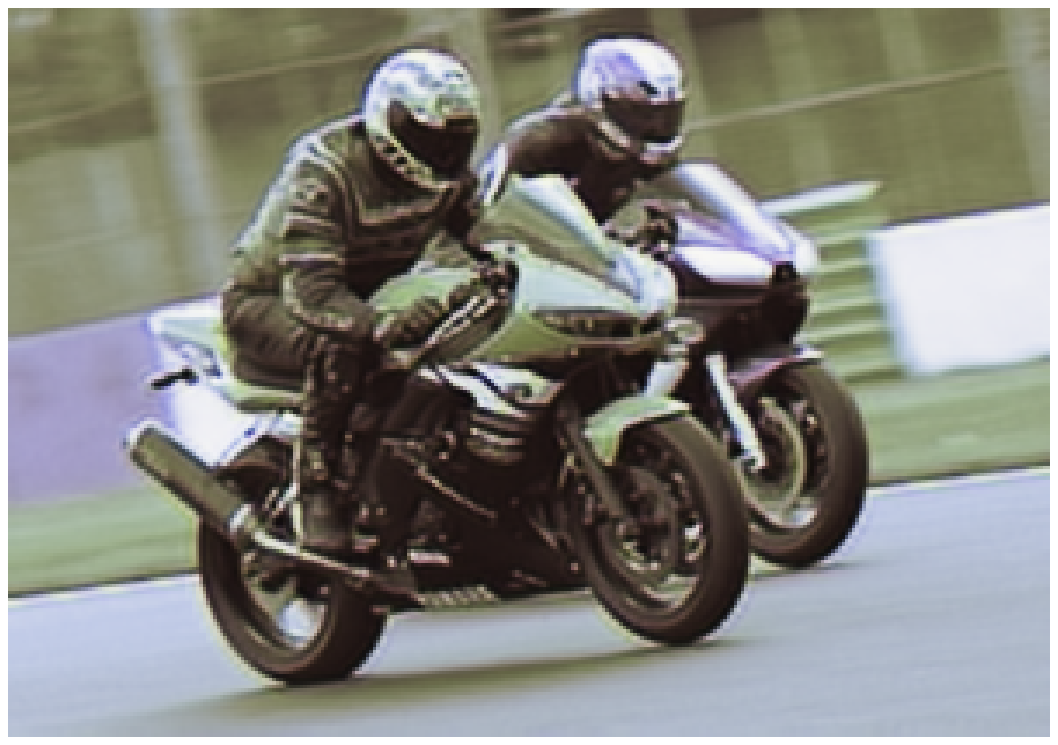}
\includegraphics[height=2.9cm]{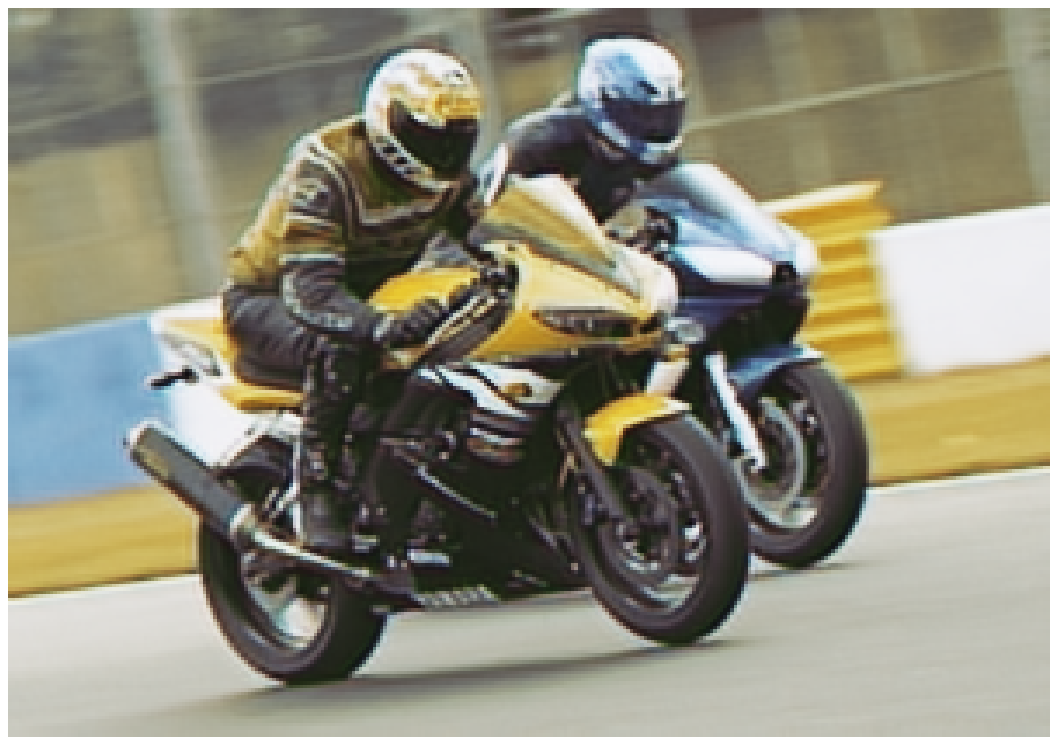}
\vspace*{-0.2cm}
\end{center}
\begin{center}
\includegraphics[height=2.9cm]{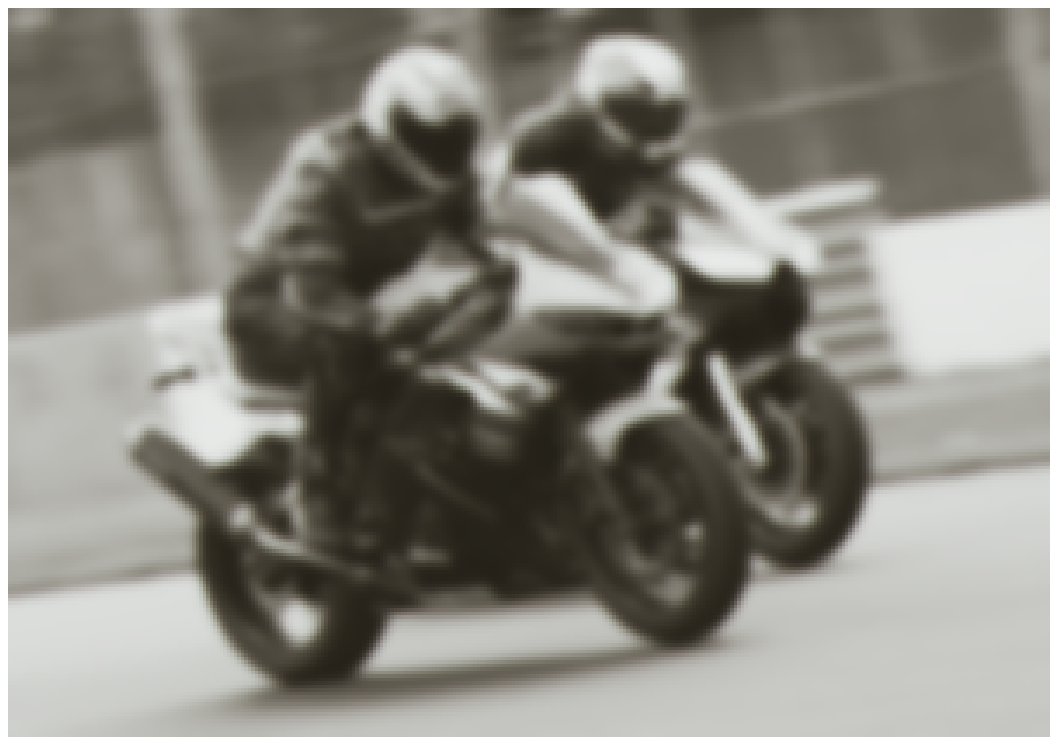}
\includegraphics[height=2.9cm]{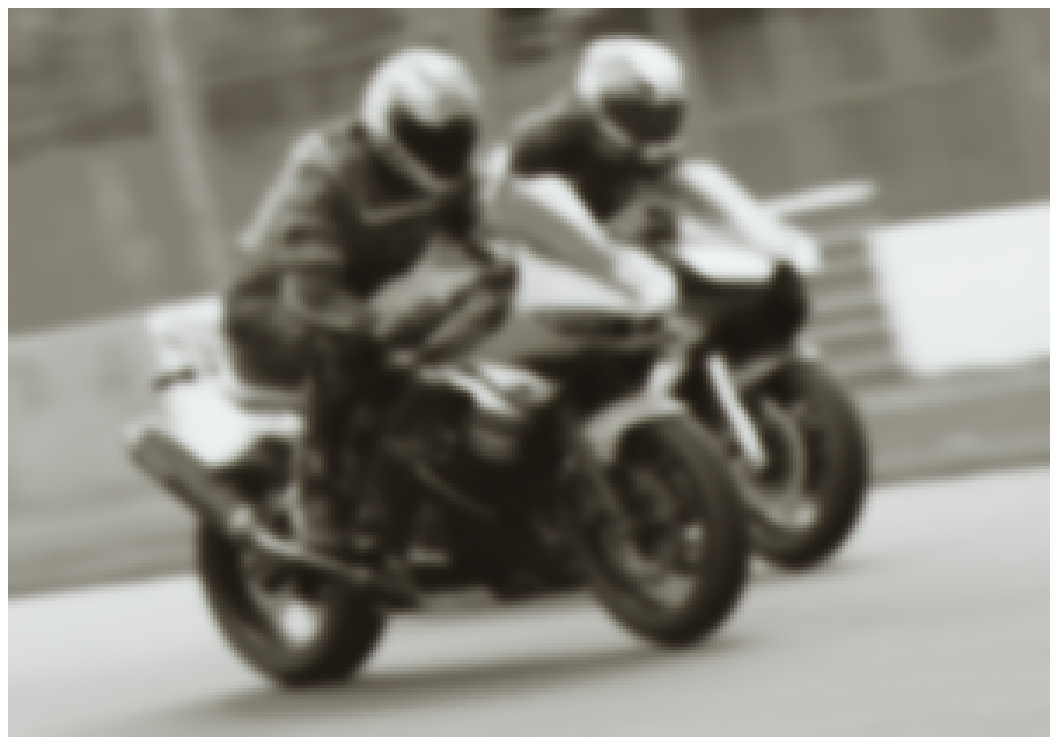}
\includegraphics[height=2.9cm]{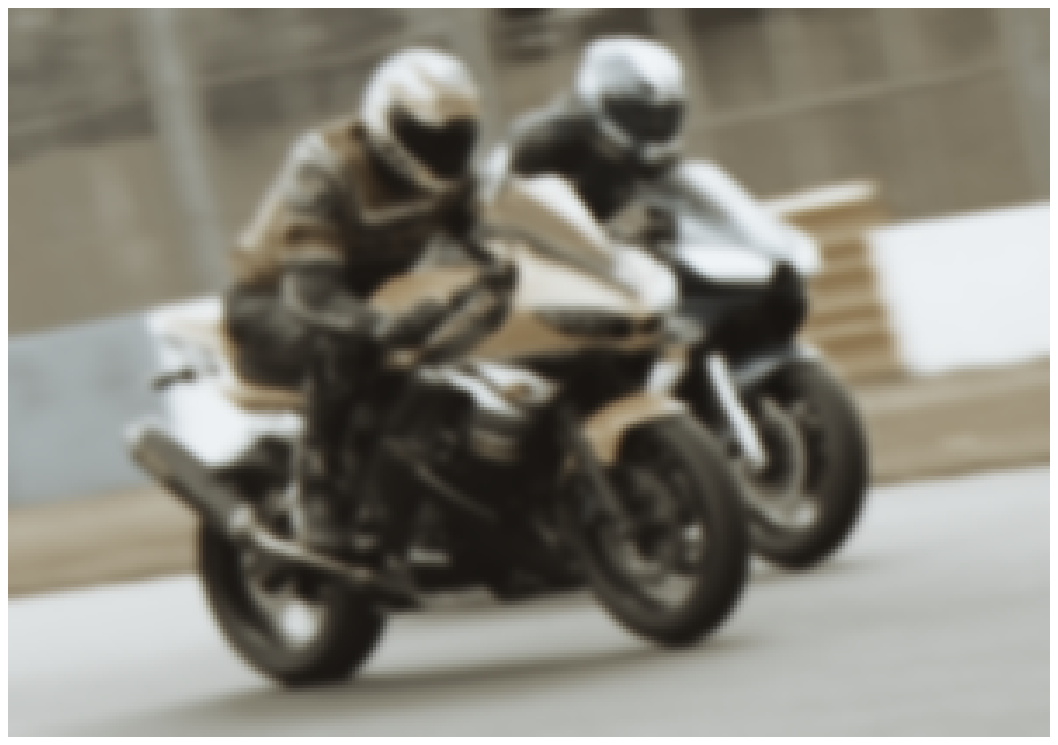}
\includegraphics[height=2.9cm]{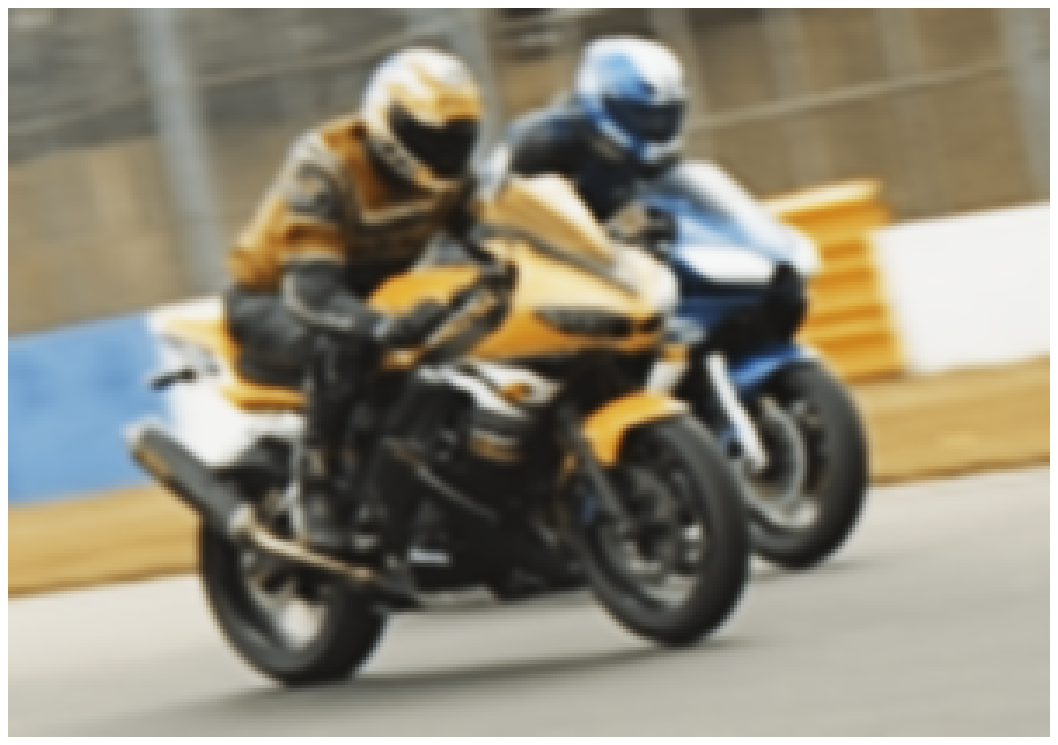}
\vspace*{-0.2cm}
\end{center}
\hspace*{8.45cm}(d)
\vspace*{0.2cm}
\caption {
\hspace{-0.2cm}(continued) The reconstructed images for the images in
Fig.\ref{fig:reconstructed_images}(c) and (d) at 100, 500, 1000 and
1500 epochs. Top row: The reconstructed images by the proposed
method. Bottom row: The reconstructed images by learning using
$E_{PL}$.
}
\label{fig:reconstructed_images_epochs}
\vspace*{-0.2cm}
\hrulefill
\vspace*{-0.3cm}
\end{figure*}
The experiments using the 70 color images in the Proposal Flow Willow
dataset\cite{Ham16} were performed to demonstrate the usefulness of
the proposed method. The size of the larger side of the original
images was equal to 300 pixels and the sizes of the other sides were
determined by the aspect ratios.\par
The architecture of the CNN for learning is shown in
Tab.\ref{tab:CNN_architecture}. From the first layer to the fifth
layer constituted a CAE. Since the activation function of the fifth
layer was $\tanh$, the pixel values of the original images were scaled
to $\left[-1,1\right]$. The resolution of the feature maps of the
third layer was the half of the resolution of the original images,
because the stride of the third layer was 2. The stride of the fourth
layer was 0.5 that means upsampling carried out by the bilinear
interpolation\cite{Shelhamer17}. The sixth layer had the weights
obtained from the Laplacian filter bank. Since the three filters with
the scales $\sigma_c=0.8,1.6$ and $3.2$ were used for the filter bank,
the sixth layer had the three channels corresponding to the
subbands. Although the size of the Laplacian filter was basically
decided by $\lceil 8\sigma_c\rceil$, the size was changed to an odd
number by adding 1 if $\lceil 8\sigma_c\rceil$ was an even number.\par
The backpropagation algorithm with the momentum term\cite{Rumelhart86}
was use for learning. The learning coefficient was 0.02 and the weight
for the momentum term was 0.5. The initial values of the filters were
random numbers given by the distribution $N\left(0,0.02\right)$ and
all the initial values of the biases were 0. All the weights
$w_{PL}\left(c\right)$ in Eq.(\ref{eq:pl}) were 1. The weights of
Eq.(\ref{eq:sfl}) were 100 for $\sigma_c=0.8$ and 10 for both
$\sigma_c=1.6$ and $3.2$, which emphasized the loss of the highest
subband. The gradients of Eq.(\ref{eq:sfl_grad}) and
(\ref{eq:pl_grad}) were computed by the
full-batch~$\left(N_m=70\right)$. The maximum epoch was 2000.\par
The learning algorithm was implemented by C++ and CUDA\cite{CUDA} and
performed on both the CPU and the GPU, Intel Xeon E5-2637v4 and NVIDIA
Quadro M4000. The computational time for an epoch was about 218 and
214 seconds for the proposed method and the learning using $E_{PL}$,
respectively. Thus learning for 2000 epochs required about 5 days.\par
Figure \ref{fig:learning_curves} shows the learning curves. The curve
in Fig.\ref{fig:learning_curves}(a) is for learning using
$E_{PL}$. The curves in Fig.\ref{fig:learning_curves}(b) represent the
changes in the SFLs of the subbands in learning of (a). Note that the
gradients of Eq.(\ref{eq:sfl_grad}) were not backpropagated in
Fig.\ref{fig:learning_curves}(b). As we can see from these figures,
the SFLs of the subbands with $\sigma_c=0.8$ and $1.6$ largely
remained even though $E_{PL}$ became small. In general, the power
spectra of images concentrate in low spatial frequencies. Thus
reducing the loss of the lowest subband was good way to reproduce the
original images and $E_{PL}$ was actually reduced quickly at the
beginning of learning. However, reducing the loss of the lowest
subband turned out to reduce the gradients of $E_{PL}$ and make the
SFLs of the other subbands, especially the SFL of the highest subband,
remain. This is the reason why the reconstructed images obtained by
learning using $E_{PL}$ have blurs.\par
Figure \ref{fig:learning_curves}(c) and (d) represent the learning
curves of the proposed method and the changes in the SFLs of the
subbands in learning of (c), respectively. Since there is one-to-many
uncertainty between the output of the Laplacian filter and the
reconstructed image, the network parameters reducing $E_{SFL}$ might
increase $E_{PL}$. Therefore, the losses in
Fig.\ref{fig:learning_curves}(c) and (d) were oscillated because the
network parameters consistent with both $E_{SFL}$ and $E_{PL}$ had to
be found out. However, the SFLs of all the subbands were reduced as
learning progressed as demonstrated in
Fig.\ref{fig:learning_curves}(d). The SFLs of the subbands at 2000
epochs were (18.9,5.69,1.37)~($\times10^{-3}$) in learning using
$E_{PL}$ and (7.08,2.65,1.34)~($\times10^{-3}$) in the proposed
method, in which the SFLs are ordered with decreasing spatial
frequencies. These results quantitatively show the usefulness of the
proposed method to reduce the SFLs of the high spatial frequency
components. Figure \ref{fig:reconstructed_images} represents the
reconstructed images for the four original images at 2000 epochs,
which were generated by the proposed method and learning using
$E_{PL}$. Figure \ref{fig:reconstructed_images_epochs} shows the
process of reconstruction for the images in
Fig.\ref{fig:reconstructed_images} at 100, 500, 1000 and 1500
epochs. The reconstructed images in these figures qualitatively
demonstrate that the blurs were clearly reduced by the proposed
method.\par
The proposed method extracts the features in subbands by a Laplacian
filter bank. By this process, we can separate the features with high
spatial frequencies from the features with low spatial
frequencies. This separation enables us to perform weighting for the
gradients of the subbands by $w_{SFL}\left(c\right)$ in
Eq.(\ref{eq:sfl_grad}) and facilitate the reconstruction of the
features with high spatial frequencies. The key to reducing blurs is
the separation of features.\par
As we can see from Fig.\ref{fig:reconstructed_images_epochs}, both
the proposed method and learning using $E_{PL}$ required a large
number of epochs to reproduce the colors, although the shapes were
reconstructed in the early stage of learning. In addition, a part of
the colors did not recall at 2000 epochs as shown in
Fig.\ref{fig:reconstructed_images}. Reproducing correct colors in a
short learning time is a problem for future work.
%
%
%
%
\section{Summary}
\label{sec:summary}
In this paper, the learning method for CAEs using the loss function
computed from features reflecting spatial frequencies has been
presented to reduce the blurs in reconstructed images. The spatial
frequency loss~(SFL) was defined using the features extracted by a
Laplacian filter bank and the learning algorithm using the loss
function constructed from the SFLs was shown. The experimental results
were given to demonstrate the usefulness of the proposed method
quantitatively and qualitatively. These results will contribute to
facilitate the use of CAEs for feature extraction.
%
%
{\small
\bibliographystyle{ieee}
\bibliography{SFL_learning_CAEs}
}
\end{document}